\pdfoutput=1

\documentclass[11pt]{article}

\usepackage[final]{acl}
\usepackage{times}
\usepackage{latexsym}
\usepackage{xcolor}         

\usepackage[dvipsnames]{xcolor}

\usepackage[T1]{fontenc}

\usepackage[utf8]{inputenc}

\usepackage{microtype}

\usepackage{inconsolata}
\usepackage{amsmath}
\usepackage{graphicx}
\usepackage[most]{tcolorbox}
\usepackage{float} 
\usepackage[capitalize]{cleveref}
\usepackage{stfloats}
\usepackage{booktabs}
\usepackage{placeins}
\usepackage{titlesec} 
\usepackage{titling} 
\usepackage{makecell}
\usepackage{booktabs}
\usepackage{subcaption}
\usepackage{pifont}
\usepackage{newunicodechar}
\newunicodechar{♫}{\ding{14}}
\usepackage{tikz}
\usepackage{enumitem}
\usepackage{sansmath}  
\usepackage{helvet}    
\usepackage[percent]{overpic} 
\crefname{figure}{Fig.}{Figs.}
\crefname{table}{Tab.}{Tabs.}
\crefname{section}{Sec.}{Secs.}
\crefname{subsection}{Sec.}{Secs.}
\crefname{subsubsection}{Sec.}{Secs.}

\definecolor{mpl_blue}{RGB}{31,119,180}
\definecolor{mpl_orange}{RGB}{255,127,14}
\definecolor{mpl_green}{RGB}{44,160,44}
\definecolor{mpl_red}{RGB}{214,39,40}
\definecolor{mpl_purple}{RGB}{148,103,189}

\title{
ReTAG: Retrieval-Enhanced, Topic-Augmented \\
Graph-Based Global Sensemaking
}

\author{%
Boyoung Kim$^{1}$ \quad Dosung Lee$^{1}$ \quad Sumin An$^{1}$ \\ \textbf{Jinseong Jeong}$^{1}$ \quad \textbf{Paul Hongsuck Seo}$^{1}$
\\
$^{1}$Dept. of CSE, 
Korea University \\
  {\small{\texttt{\{bykimby, dslee1219, suminan, dw9030, phseo\}@korea.ac.kr}}} \\
}

\newcommand{\eg}{\textit{e.g.}}

\begin{document}
\maketitle
\begin{abstract}
Recent advances in question answering have led to substantial progress in tasks such as multi-hop reasoning.
However, global sensemaking—answering
questions by synthesizing information 
from
an entire corpus—remains a significant challenge. 
A prior graph-based approach to global sensemaking lacks retrieval mechanisms, topic specificity, and incurs high inference costs. To address these limitations, we propose ReTAG, a Retrieval-Enhanced, Topic-Augmented Graph framework that constructs topic-specific subgraphs and retrieves the relevant summaries for response generation. Experiments show that ReTAG improves response quality while significantly reducing inference time compared to the baseline. Our code is available at \href{https://github.com/bykimby/retag}
{https://github.com/bykimby/retag}.

\end{abstract}

\section{Introduction}
\begin{figure}[!t]
  \centering  
  \includegraphics[width=\columnwidth]{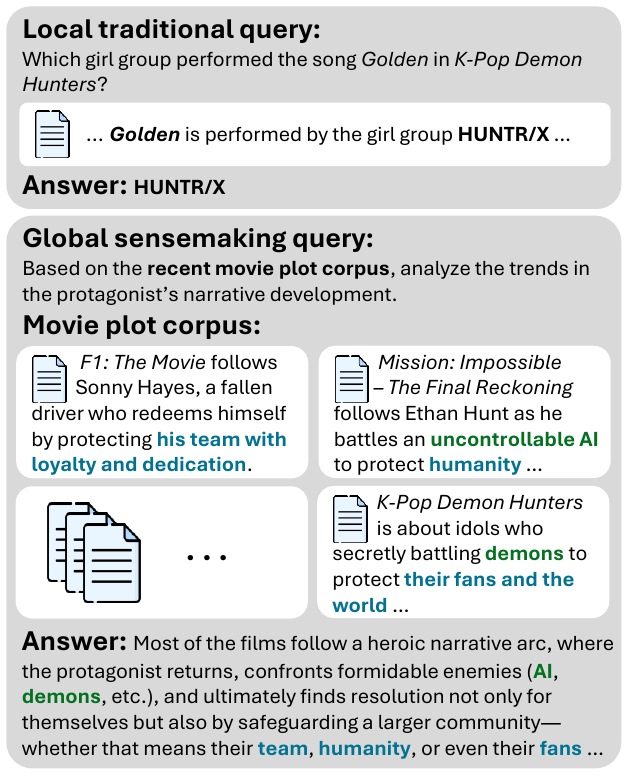}
  \caption{ \textbf{Comparison between Local Traditional QA and Global Sensemaking QA.} 
  Local traditional QA focuses on local details in documents (the performer of \textit{Golden}), while global sensemaking QA requires examining entity relationships (AI, humanity, etc.) across the entire corpus.
  }
  \label{fig:task}
\end{figure}

Global sensemaking~\cite{edge2025localglobalgraphrag} refers to the task of synthesizing comprehensive information across a large corpus of documents to generate a coherent response to a query. 
For instance, consider the global sensemaking query in \cref{fig:task}: `\textit{Based on the recent movie plot corpus, analyze the trends in the protagonist’s narrative development.}' 
Answering this question requires examining not only character–event relations within a single movie plot document but also shared and divergent narrative patterns across movie plot corpus. Such queries require analyzing the interdependencies among diverse entities (e.g., characters, events) across the entire corpus, which makes global sensemaking a highly challenging task. In contrast, local traditional QA targets fact-based queries with specific documents, such as identifying which girl group performed \textit{Golden} in \textit{K-Pop Demon Hunters} in \cref{fig:task} \citep{rajpurkar2016squad, kwiatkowski2019natural, yao2023react, lee2025rescore}. Unlike such localized approaches, global sensemaking requires integrating a broader range of knowledge, often spanning across multiple sources.


The task of global sensemaking presents several challenges.
A key issue is that the vast amount of information within the corpus $\mathcal{D}$ cannot be directly processed by a large language model (LLM) due to its limited context window~\cite{wang2024beyond}. 
Identifying the relevant entities and relations for a given query and representing their complex inter-dependencies in a way that is both efficient and meaningful for the LLM is an ongoing challenge. 

A recent baseline approach for global sensemaking, proposed by \citet{edge2025localglobalgraphrag}, uses a graph-based method that captures the relationships among entities across the entire corpus.
The method constructs a contextualized entity-relation graph and generates community summaries by clustering related subgraphs. 
While effective, this method relies on a single general set of summaries for all queries, making it less flexible and potentially less efficient for capturing specific details or nuances in global sensemaking tasks. 
Additionally, it struggles with handling large document corpora due to the high complexity of processing the entire graph.

To address these challenges, we propose ReTAG, a Retrieval-Enhanced, Topic-Augmented Graph-based approach for global sensemaking. 
ReTAG builds upon the baseline graph-based approach by incorporating topic-augmented summarization and retrieval augmentation techniques to provide a more efficient and effective response generation process.
Through our experiments, we demonstrate that ReTAG not only improves the quality of the responses but also significantly reduces inference time, making it a robust solution for global sensemaking in large-scale document corpora. Our contributions are three-fold:
\begin{itemize}[itemsep=2pt, topsep=0pt, parsep=0pt, partopsep=0pt]
    \item We introduce topic-augmented community summarization, reducing fragmentation and enhancing relevance for global sensemaking.
    \item We propose keyword-expanded retrieval augmentation, selecting the relevant community summaries to significantly improve efficiency.
    \item We conduct extensive experiments demonstrating the effectiveness of our framework across datasets, improving both response quality and inference time.
\end{itemize}
    
\section{Related Work}

\noindent \textbf{Global Sensemaking} \ \ 
Early multi-document summarization methods focused on generating concise summaries from document sets~\cite{radev2004centroid, erkan2004lexrank}, while recent work has explored modeling inter-document relationships using neural topic models~\cite{bianchi2020pre, cui2021topic} and graph-based representations~\cite{yasunaga2017graph, li2020leveraging, wu2021bass}. 
These studies focus on summarization but do not extend to question answering. \citet{edge2025localglobalgraphrag} advances global sensemaking for QA but uses a single general set of  summaries for all queries.
In contrast, we enhance global sensemaking by incorporating topic-augmented summarization and retrieval-based filtering, improving both effectiveness and efficiency.


\noindent \textbf{Retrieval-Augmented QA} \ \ 
QA research has focused on retrieval-augmented approaches that identify the most relevant document for a given query, evaluated across various benchmark datasets~\cite{rajpurkar2016squad, yang2018hotpotqa, kwiatkowski2019natural, ho2020constructing, trivedi2022musique}. 
Methods like REALM~\cite{guu2020retrieval}, RAG~\cite{lewis2020retrieval}, and Self-RAG~\cite{asai2023self} use retrievers to find a closely aligned document and generate answers based on it. 
Building on this, approaches such as IRCoT~\cite{trivedi2022interleaving}, ReAct~\cite{yao2023react}, and ReSCORE~\cite{lee2025rescore} extend the retrieval process to multiple documents for more complex reasoning. 
However, these methods focus on subsets of the corpus and lack the ability to capture global sense, which is the focus of this work.


\noindent \textbf{Graph-Based QA} \ \ 
A line of research has explored graph-based QA approaches that model relations among entities across documents. 
Several studies~\cite{asai2019learning, sun2023think, wang2024knowledge} leverage graph structures to identify reasoning paths and retrieve relevant context for QA.
KG-GPT~\cite{kim2023kg}, G-Retriever~\cite{he2024g} and GRAG~\cite{hu2024grag} perform reasoning over query-focused subgraphs constructed from retrieved graph elements. 
However, graph-based QA approaches for global sensemaking have yet to be fully explored. 
\citet{edge2025localglobalgraphrag} addresses this challenge by building a graph that captures the context of the entire corpus.
Our work extends this method by incorporating topic-augmented graph construction and community retrieval, enhancing both response quality and efficiency.

\section{Methods}

\begin{figure}[t]
  \centering  
  \includegraphics[width=0.85\columnwidth]{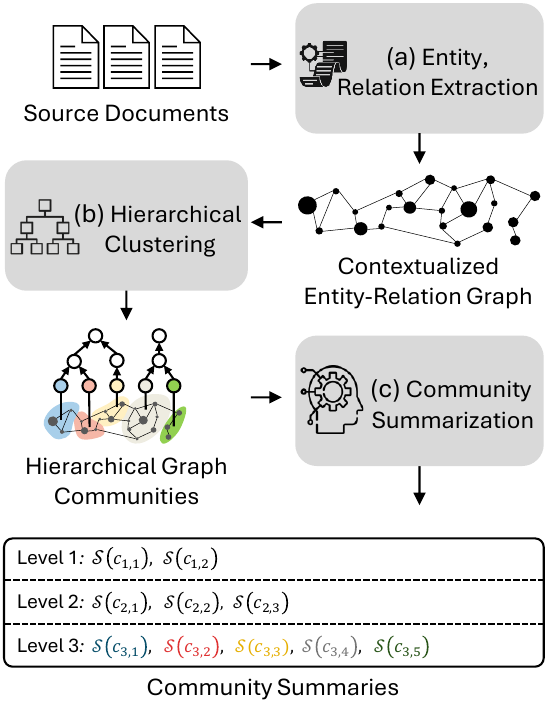}
  \caption{
  \textbf{Overall Process of Graph-based Global Sensemaking.} All entities and relations are extracted by an LLM to build a contextualized entity-relation graph (a), which is then clustered forming hierarchical communities (b).
  A community summary is generated for each community based on the contexts within its corresponding subgraph (c).
  }
  \label{fig:document}
\end{figure}

Given a query $q$ and a document database $\mathcal{D}$, global sensemaking is defined as the task of generating a response $a$ that synthesizes comprehensive information from the entire corpus of documents $\mathcal{D}$. 
This task involves identifying the relations among entities (people, places, and events, etc) relevant to 
the given query $q$ within the entire corpus $\mathcal{D}$, and integrating this information to produce a coherent and informed response. 
This task contrasts to existing QA and multi-hop QA tasks, performing detail-oriented QA on a subset of documents.
This characteristic poses a significant challenge, as it is practically infeasible to incorporate $\mathcal{D}$ into 
the limited context window of LLMs. 
Furthermore, identifying the entities and relations pertinent to the query $q$ from $\mathcal{D}$, and conveying their complex inter-dependencies in a textual form that the LLM can efficiently process remains a significant and ongoing challenge.

To address these challenges, we propose \textbf{ReTAG}, a Retrieval-Enhanced, Topic-Augmented Graph-based global sensemaking approach.
Building on the prior work of graph-based global sensemaking \citep{edge2025localglobalgraphrag}, which is outlined in \cref{subsec:graph-based-global-sensemaking}, we introduce topic and retrieval augmentation techniques in \cref{subsec:topic-augmentation} and \cref{subsec:retrieval-augmentation}, respectively.
Finally, \cref{subsec:retag} presents ReTAG, our complete framework that integrates both techniques.

\subsection{Graph-based Global Sensemaking}\label{subsec:graph-based-global-sensemaking}
\citet{edge2025localglobalgraphrag} recently proposed a graph-based approach for global sensemaking. 
The method begins by constructing a contextualized entity-relation graph (\cref{subsubsec:contextualized-entity-relation-graph}), which systematically organizes the interdependencies between entities and relations across the entire document database $\mathcal{D}$, while preserving the contexts in which they are mentioned. 
This graph is then hierarchically clustered, grouping entities and relations with shared thematic content into distinct communities. 
Subsequently, a community summary is generated for each community, drawing from the contexts linked to the subgraph within the community (\cref{subsubsec:community-based-summarization}). 
These summaries are utilized to produce a comprehensive response $a$ to the given query $q$ (\cref{subsubsec:response-generation}).

\subsubsection{Contextualized Entity-Relation Graph} \label{subsubsec:contextualized-entity-relation-graph}
To enable effective global sensemaking, it is essential to gather pertinent information distributed across the entire document corpus $\mathcal{D}$. 
Specifically, a contextualized entity-relation graph $G_c=(\mathcal{V}, \mathcal{E})$ is constructed where entities are represented as nodes $n_i\in \mathcal{V}$ and relations as edges $e_i \in \mathcal{E}$. 
Each entity $n_i$ is associated with a set of relevant contexts $\mathrm{ctx}(n_i)$, which consists of sentences containing the entity, collected from across $\mathcal{D}$.
Similarly, each relation $e_i$ between two entities is paired with a set of contexts $\mathrm{ctx}(e_i)$, representing sentences in $\mathcal{D}$ containing the relation.
This contextualized graph captures the structure of $\mathcal{D}$ by collecting relevant statements anchored by entities or relations.
For the graph construction, an LLM is prompted to simultaneously extract entities, relations, and their corresponding contexts from each document $d \in \mathcal{D}$. 
The graph $G_c$ is then formed by merging the extracted entities and relations. 
For a detailed description of the graph construction process, refer to \cref{appendix:lowest-level-graph}.

\subsubsection{Community-Based Summarization}\label{subsubsec:community-based-summarization}
While the graph $G_c$ organizes the corpus $\mathcal{D}$ to capture global information for different entities and relations, it is impractical to prompt the LLM with the entire graph as context. 
To address this, \citet{edge2025localglobalgraphrag} proposes forming communities by clustering subgraphs 
using an agglomerative clustering algorithm, subsequently generating a report-like summary for each community to be used in answer generation.

In particular, the Hierarchical Leiden algorithm~\cite{DBLP:journals/corr/abs-1810-08473} is applied to $G_c$, resulting in $L$ multi-level sets of communities $\mathcal{C}_{l} = {c_{l,i}}_{i=1}^{N_l}$, where $c_{l,i}$ represents a community at level $l$, with level $1$ being the highest and level $L$ being the lowest (leaf) level.
The set of communities $\mathcal{C}_l$ at each level $l$ partitions the entire graph.
It is also worth mentioning that the algorithm does not require all $L$ sets of communities for global sensemaking, but just one. However, at different levels, the granularity of the partitioning varies. 
Therefore, the quality of responses is tested at different granularities, as in \citet{edge2025localglobalgraphrag}, while maintaining all $L$ communities.

Given the constructed hierarchical communities, a community summary is generated by consolidating the contexts associated with the entities and relations within the graph.
Specifically, to generate a summary, a community context $\mathrm{ctx}(c_{l,i})$ is first constructed by collecting all contexts linked to the nodes and edges within the community, namely $\mathrm{ctx}(n_j)$ and $\mathrm{ctx}(e_j)$.
An LLM is then prompted with this community context to generate a summary $\mathcal{S}(c_{l,i})$.
For a detailed description of the summarization process, refer to \cref{appendix:hierarchical-graph-summaries}.

\subsubsection{Response Generation}\label{subsubsec:response-generation}

The final global sensemaking response is generated by aggregating community summaries using an LLM. 
For a community set $\mathcal{C}_{l}$ at level $l$, the communities $c_l \in \mathcal{C}_{l}$ are shuffled and sequentially prompted to generate sub-answers until the token limit is reached. 
This process is repeated until all summaries are consumed, yielding a set of sub-answers $a_{l,j}$. 
These are then aggregated into a global answer $a_l$, which represents the entire corpus $\mathcal{D}$ through the partitioned community set $\mathcal{C}_l$ at level $l$. 
The hierarchical community structure allows for generating global answers at different levels, balancing summary detail and global coverage. For a detailed description of the response process, refer to \cref{appendix:response-generation}.

\subsubsection{Limitations} \label{subsubsec:limitations}
While this graph-based approach effectively generates global sensemaking responses leveraging a contextualized entity-relation graph $G_c$, it has several limitations. 
A single general graph represents the entire corpus, requiring each community to cover a broad range of information, which can lead to overflow and missed details. 
Summarizing across all topics increases the risk of overlooking important content in favor of general information, potentially failing to address specific query aspects.
Additionally, communities at larger level $l$ may fragment relevant information, complicating global sensemaking.
Moreover, the framework generates sub-answers that cover the entire graph, even when communities are loosely related, introducing significant computational complexity.

\begin{figure}[t]
  \centering
  \includegraphics[width=\columnwidth]{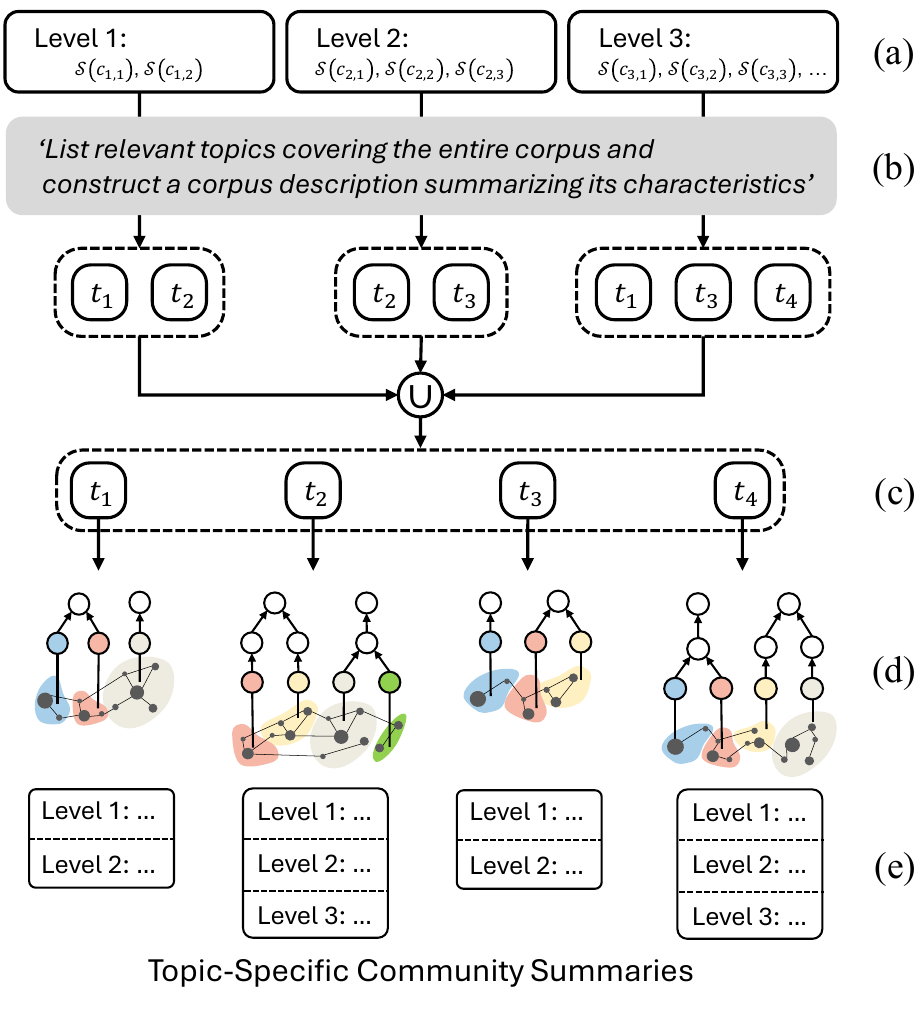}
  \caption{\textbf{Topic Augmentation Pipeline.}
  At each level $l$, given a set of community summaries (a), 
  mining all relevant topics is essentially equivalent to performing global sensemaking on the given summaries (b).
  The final set of topics is constructed by taking the union of the sets of topics across all levels (c).
  For each topic, we apply the same community summary generation process (d), resulting in multi-level sets of community summaries generated for each topic (e).
  }
  \label{fig:tagging-pipeline}
\end{figure}

\subsection{Topic Augmentation}\label{subsec:topic-augmentation}

To address the limitations outlined, we propose topic augmentation for graph-based global sensemaking.
This technique focuses on maintaining essential content by constructing separate community sets around specific target topics, reducing content fragmentation and strengthening graph connections.
It prioritizes eliminating redundant information over discarding unique content, preserving crucial details for global sensemaking.
Additionally, the smaller graph size and fewer community summaries per topic reduce computational complexity during inference.

In \cref{subsubsec:topic-augmented community-summarization}, we first discuss topic-augmented community summarization, building on the general community summaries from \cref{subsubsec:community-based-summarization}, followed by the topic-augmented response generation procedure in \cref{subsubsec:topic-augmented-response-generation}.

\subsubsection{Topic-Augmented Community Summarization}
\label{subsubsec:topic-augmented community-summarization}

Our goal is to propose an algorithm that generates separate sets of community summaries, each covering the entire corpus $\mathcal{D}$ while focusing on a specific topic.
We first mine related topics likely to span $\mathcal{D}$ and use these to construct topic-augmented community summaries.

\noindent \textbf{Comprehensive Topic Mining} \ \ 
We begin by mining a set of topics $\mathcal{T}$ that comprehensively cover global sensemaking queries for $\mathcal{D}$, focusing on those likely to be asked. 
This task is framed as a global sensemaking query for $\mathcal{D}$.
Specifically, we form global sensemaking queries: `list relevant topics covering the entire corpus' and `construct a corpus description summarizing its characteristics.' 
The response is generated through the process outlined in \cref{subsubsec:response-generation}.
We query at all $L$ levels, obtaining $L$ sets of topics as answers, the union of which forms the final set of extracted topics $\mathcal{T}$.
Note that the additional dataset description generated is used later during inference, as described in \cref{subsubsec:topic-augmented-response-generation}.

\noindent \textbf{Topic-Augmented Summary Generation} \ \ 
Given a set of extracted topics $\mathcal{T}$, we construct a separate hierarchical community-based summarization for each topic $t \in \mathcal{T}$. 
To construct the topic-augmented graph $G_c^{(t)}$, we extract only the entities and relations related to the target topic $t$ from the document corpus.
This 
narrows the graph to a topic-augmented subset, reducing inference time and improving answer quality by excluding irrelevant information.
Once we have $G_c^{(t)}$, we apply the procedure in \cref{subsubsec:community-based-summarization} to obtain topic-augmented communities $c_{l}^{(t)} \in \mathcal{C}_l^{(t)}$ and their corresponding summaries $\mathcal{S}(c_{l}^{(t)})$, which are used for topic-augmented response generation.
\subsubsection{Topic-Augmented Response Generation}
\label{subsubsec:topic-augmented-response-generation}
Once topic-augmented community summaries are ready, we identify the target topic $t$ and generate the response $a$ to the given query $q$ based on the topic-augmented summaries.

\noindent \textbf{Topic Classification} \ \
To identify the target topic $t$ for the query $q$, we prompt the LLM to classify $q$ within the set of topics $\mathcal{T}$. 
Since the same terms may refer to different topics depending on the domain, we include a corpus description $\mathcal{D}$ in the prompt to assist the LLM in inferring the correct topic. 
Note that this description is generated during the topic mining process in \cref{subsubsec:topic-augmented community-summarization}. 
If no appropriate topic is found, the LLM returns null.
Related prompts can be found in \cref{appendix:prompts}.

\noindent \textbf{Response Generation} \ \ 
Once the target topic $t$ is identified, we generate the response using the topic-augmented community summaries for $\mathcal{C}^{(t)}_l$ at level $l$, while following the process outlined in \cref{subsubsec:response-generation}. 
If the LLM fails to classify $q$ into a topic, the general summaries based on $\mathcal{C}_l$ are used instead.

\subsection{Retrieval Augmentation}
\label{subsec:retrieval-augmentation}
Inefficiency in inferring global sensemaking responses arises from the large graph $G_c$ covering the entire corpus $\mathcal{D}$. 
Topic-augmented community summarization reduces the size of $G_c^{(t)}$ for each topic, but each topic-augmented graph can still be large. 
To improve efficiency, we incorporate retrieval into the process.
Given a graph $G_c$ (which may be topic-augmented), the query $q$, and the level $l$, we retrieve the top $p$ most relevant community summaries $\mathcal{S}(c_l)$. 
Only these summaries are used to generate the response using the regular process described in \cref{subsubsec:response-generation}. To enhance retrieval quality, we introduce a keyword expansion technique, prompting the LLM to generate keywords useful for retrieving relevant documents based on $q$ and the document description from \cref{subsubsec:topic-augmented community-summarization}. 
This keyword expansion (\cref{prompt:keyword-expansion}) produces keywords that are both corpus-relevant and effective for retrieval.

\subsection{ReTAG}\label{subsec:retag}
By integrating topic and retrieval augmentation techniques, we introduce ReTAG, a framework for graph-based global sensemaking.
ReTAG enhances response quality and efficiency by narrowing the scope of each topic-augmented graph, ensuring more focused answers.
Retrieval augmentation further optimizes the process by selecting the most relevant community summaries, speeding up inference. 
This combination makes ReTAG a robust and efficient solution for global sensemaking.
\section{Experiments}
\subsection{Experiment Setup}
\subsubsection{Document Corpora}

In this study, we follow the experimental setup of 
\citet{edge2025localglobalgraphrag} and utilize two corpora:

\noindent \textbf{Podcast} \citep{edge2025localglobalgraphrag} \ \ 
Podcast is constructed from publicly available transcripts of the `\textit{Behind the Tech with Kevin Scott},' presenting dialogues with leading experts across science and technology fields. This corpus is divided into $1634\times600$-token chunks, with 100-token overlaps.

\noindent\textbf{News Articles} \citep{tang2024multihopragbenchmarkingretrievalaugmentedgeneration} \ \ 
News Articles comprises multiple real-world domains, including entertainment, business, sports, technology, health, and science, thus reflecting a broad range of informational contexts. This corpus is divided into $3169\times600$-token chunks, with 100-token overlap.

\subsubsection{Global Sensemaking Queries}
Following \citet{edge2025localglobalgraphrag}, we reproduced the queries for each corpus using the \texttt{GPT-4o} API. The queries were designed to encourage comprehensive answers that reflect the entire document corpus. 
To generate the queries, we provided the LLM with high-level descriptions of each corpus, extracted from the corresponding publication or official website. Based on the description, the LLM was prompted to generate global sensemaking questions by jointly creating hypothetical users and tasks (See \cref{appendix:query} for details).

\subsubsection{Models}
We re-implemented a baseline model based on \citet{edge2025localglobalgraphrag}, as described in \cref{subsec:graph-based-global-sensemaking}, and compare ReTAG with it.
Both are implemented using the Llama 3.3 70B Instruct FP8 model.
We use BM25 \citep{robertson1995okapi} for retrieval augmentation and run the LLM with a 4K context token limit.

\begin{table}[t]
\centering
\scalebox{0.75}{
\begin{tabular}{lccccc}
\toprule
& \multicolumn{2}{c}{Podcast}
& \multicolumn{2}{c}{News Articles}\\
\cmidrule(lr){2-3}\cmidrule(lr){4-5}

& w/o TA & w/ TA
& w/o TA & w/ TA\\
\midrule
\# nodes & 5,021   & 3,277 & 18,673    & 13,626\\
\# edges & 10,658  & 6,953 & 35,350    & 22,288 \\
\bottomrule
\end{tabular}
}
\vspace{-0.3em} 
\caption{\textbf{The Node and Edge Counts in Models with and without Topic Augmentation (TA).} Comparisons of node and edge counts in the contextualized entity-relation graphs. 
Note that rounded average across topics is reported for the one with TA.}
\label{tab:graph_stats}
\end{table}

\begin{table}[t]
\centering
\scalebox{0.75}{
\begin{tabular}{ccccc}
\toprule
& \multicolumn{2}{c}{Podcast}
& \multicolumn{2}{c}{News Articles}\\
\cmidrule(lr){2-3}\cmidrule(lr){4-5}
Level
& w/o TA
& w/ TA
& w/o TA
& w/ TA\\
\midrule
1 &  29 &  31&    60 &   72 \\
2 & 306 & 228&   716 &  655 \\
3 & 668 & 473 & 2,251 & 1,693 \\
4 & 773 & 534 & 2,987 & 2,042\\
5 & 786 & 545 & 3,110 & 2,093 \\
6 & 787 & 578 & 3,126 & 2,100\\
7 & -- & -- & 3,129 & 2,067  \\
\bottomrule
\end{tabular}}
\vspace{-0.3em} 
\caption{
\textbf{Community Summary Counts in Models with and without Topic Augmentation (TA).}
Number of community summaries used per query at each level for News Articles and Podcast. Rounded average across topics used for the model with TA.}
\label{tab:community_per_question}
\end{table}

\subsubsection{Evaluation Criteria}
To assess global sensemaking performance, we adopt the LLM based winning rates with head-to-head comparison from \citet{edge2025localglobalgraphrag} (See prompt in \cref{appendix:prompts}).
It involves prompting \texttt{GPT-4o-mini} to judge which of the responses is better.
We adopt three evaluation criteria: 

\noindent\textbf{Comprehensiveness} \ \ 
This evaluates how rich and detailed the response content is. 

\noindent\textbf{Diversity} \ \  This evaluates how well the response incorporates diverse perspectives or information.

\noindent\textbf{Empowerment} \ \ This evaluates how much the response provides practical value to the user.

The target model is compared to the baseline, with a winning rate above 50\% indicating that the target model outperforms the baseline.
We exclude \textit{Directness} based on its negative correlation with global sensemaking reported in \citet{edge2025localglobalgraphrag}.
In addition, we also report the inference time per question, calculated by dividing the total inference time for answering all questions in a batch (with a batch size of 256) by the number of questions using two NVIDIA H100 GPUs.

\subsection{Effects of Topic Augmentation}\label{subsec:exp-topic-augmentation}
\noindent\textbf{Topics Mined} \ \ 
From the topic mining process introduced in \cref{subsubsec:topic-augmented community-summarization}, we identified 29 and 38 topics for News Articles and Podcast, respectively, as detailed in \cref{appendix:topic-qualitative-analysis}.
These topics provide broad coverage of the entire corpus (\eg, \textit{Artificial Intelligence}, \textit{Mathematics}, \textit{Software Development} in Podcast) while capturing all relevant sub-areas, enabling effective handling of diverse questions across different topics.

\noindent\textbf{Graph and Community Set Size} \ \ 
\cref{tab:graph_stats} presents the number of nodes and edges in the contextualized entity-relation graph $G_c$ for the baseline with and without topic augmentation across both datasets.
In the case of topic augmentation, the numbers represent the averages across topics.
Topic augmentation achieves average reduction of 30.88\% in nodes and 35.86\% in edges across datasets.
This reduction results in fewer community summaries at each level on average, which directly impacts inference efficiency, as shown in \cref{tab:community_per_question}.

\begin{table}[t]
\centering
\scalebox{0.75}{
\begin{tabular}{ccccc}
\toprule
& \multicolumn{2}{c}{Podcast}
& \multicolumn{2}{c}{News Articles}\\
\cmidrule(lr){2-3}\cmidrule(lr){4-5}
Level
& w/o TA
& w/ TA
& w/o TA
& w/ TA\\
\midrule
1 & 10 \small{(34.5\%)} & 19 \small{(61.3\%)} 
& 2 \small{(3.3\%)} &  7 \small{(9.7\%)} \\
2 & 34 \small{(11.1\%)} & 87 \small{(38.2\%)} 
& 12 \small{(1.7\%)} & 26 \small{(4.0\%)} \\
3 & 75 \small{(11.2\%)} & 180 \small{(38.1\%)} 
& 31 \small{(1.4\%)} & 59 \small{(3.5\%)} \\
4 & 90 \small{(11.6\%)} & 207 \small{(38.8\%)} 
& 38 \small{(1.3\%)} & 72 \small{(3.5\%)} \\
5 & 91 \small{(11.6\%)} & 210 \small{(38.5\%)} 
& 39 \small{(1.3\%)} & 74 \small{(3.5\%)} \\
6 & 91 \small{(11.6\%)} & 212 \small{(36.7\%)} 
& 39 \small{(1.2\%)} & 74 \small{(3.5\%)} \\
7 & -- & -- &  39 \small{(1.2\%)} & 55 \small{(2.7\%)} \\
\bottomrule
\end{tabular}}
\vspace{-0.3em} 
\caption{\textbf{Average Number of Relevant Summaries Identified by LLM for Each Query.}
We compare models with and without topic augmentation (TA). For the one with TA, the average is reported across topics.
}
\label{tab:llama-gt-community}
\end{table}

\noindent\textbf{Summary-to-Query Relevance} \ \ 
We also evaluate the relevance of each community summary to the input query based on LLM prompting. 
\cref{tab:llama-gt-community} presents the average proportion of relevant summaries identified.
The baseline method shows only 1 to 3\% of summaries as relevant to the input query on News Articles, whereas topic augmentation more than doubles this proportion. 
Similar trend is observed in Podcast on a larger scale.
The significantly higher proportion of relevant summaries clearly demonstrates that topic augmentation enables each graph to retain details pertinent to the specific query, leading to richer and more focused responses.

\begin{table}[t]
\centering
\begin{subtable}[t]{\linewidth}
\centering
\begin{subtable}[t]{\linewidth}
\centering
\scalebox{0.75}{
\begin{tabular}{cccccc}
\toprule
\multicolumn{1}{c}{}
& \multicolumn{3}{c}{Winning Rate (\%) $\uparrow$}
& \multicolumn{2}{c}{Inference Time (sec) $\downarrow$} 
\\
\cmidrule(lr){2-4}
\cmidrule(lr){5-6}
Level & Comp. & Div. & Emp. & w/o TA & w/ TA \\
\midrule
1 & 76.4 & 56.0 & 68.4 & 1.46 & 1.49 \\
2 & 82.0 & 73.6 & 79.6 & 8.71 & 8.15 \\
3 & 71.6 & 67.2 & 80.4 & 17.58 & 16.60 \\
4 & 74.0 & 67.2 & 75.2 & 20.04 & 19.09 \\
5 & 78.0 & 64.4 & 75.6 & 20.33 & 19.27 \\
6 & 73.0 & 67.7 & 78.2 & 20.28 & 19.32 \\
\bottomrule
\end{tabular}
}
\vspace{-0.3em} 
\caption*{(a) Podcast}
\end{subtable}

\vspace{0.5em} 

\scalebox{0.75}{
\begin{tabular}{cccccc}
\toprule
\multicolumn{1}{c}{}
& \multicolumn{3}{c}{Winning Rate (\%) $\uparrow$}
& \multicolumn{2}{c}{ Inference Time (sec) $\downarrow$}\\
\cmidrule(lr){2-4}
\cmidrule(lr){5-6}
Level & Comp. & Div. & Emp. & w/o TA & w/ TA \\
\midrule
1 & 58.8 & 57.6 & 54.0 & 2.36 & 2.68  \\
2 & 62.4 & 64.8 & 63.2 & 20.93 & 20.06  \\
3 & 60.0 & 55.2 & 59.2 & 58.64 & 49.56  \\
4 & 61.2 & 56.4 & 56.4 & 75.40 & 60.52  \\
5 & 56.8 & 55.6 & 62.8 & 80.97 & 62.12  \\
6 & 63.6 & 60.8 & 62.4 & 77.96 & 62.36  \\
7 & 58.8 & 59.3 & 58.2 & 77.96 & 62.54 \\
\bottomrule
\end{tabular}}
\vspace{-0.3em} 
\caption*{(b) News Articles}
\end{subtable}
\caption{\textbf{Effects of Topic Augmentation.} We measured the winning rates (\%) of the model with topic augmentation (TA). Inference time is reported as well. The results show that TA is both effective and efficient. }
\label{tab:topic_graph_main}
\end{table}
  
\noindent\textbf{Performance and Inference Time} \ \ 
We compare the performances of the baseline with and without topic augmentation in \cref{tab:topic_graph_main}, which reports winning rates and the inference time.
We observe that, with topic augmentation, the method generally outperforms the one without it. 
This improvement is attributed to the topic-specific graphs and community summaries, which enable more condensed and focused information organization, leading to higher-quality responses that are more relevant to the input query.
Additionally, the smaller topic-specific graphs facilitate faster inference, as the number of LLM prompts required is reduced accordingly. 
In this case, the inference time shows greater improvements for News Articles compared to Podcast-for example, at Level 6 of News Articles, inference time was reduced by 20.01\%, whereas in Podcast it decreased by 4.73\%. As explained in \cref{appendix:topic-qualitative-analysis}, News Articles cover a broader domain than Podcast, which enables more effective content filtering. Consequently, this reduces the graph size in News Articles and leads to substantial efficiency gains. The statistical significance of \cref{tab:topic_graph_main} was reported in \cref{app:llm-eval}.

\begin{figure}[t]
  \centering
  \begin{subfigure}[t]{\linewidth}
    \centering
    \includegraphics[width=\linewidth]{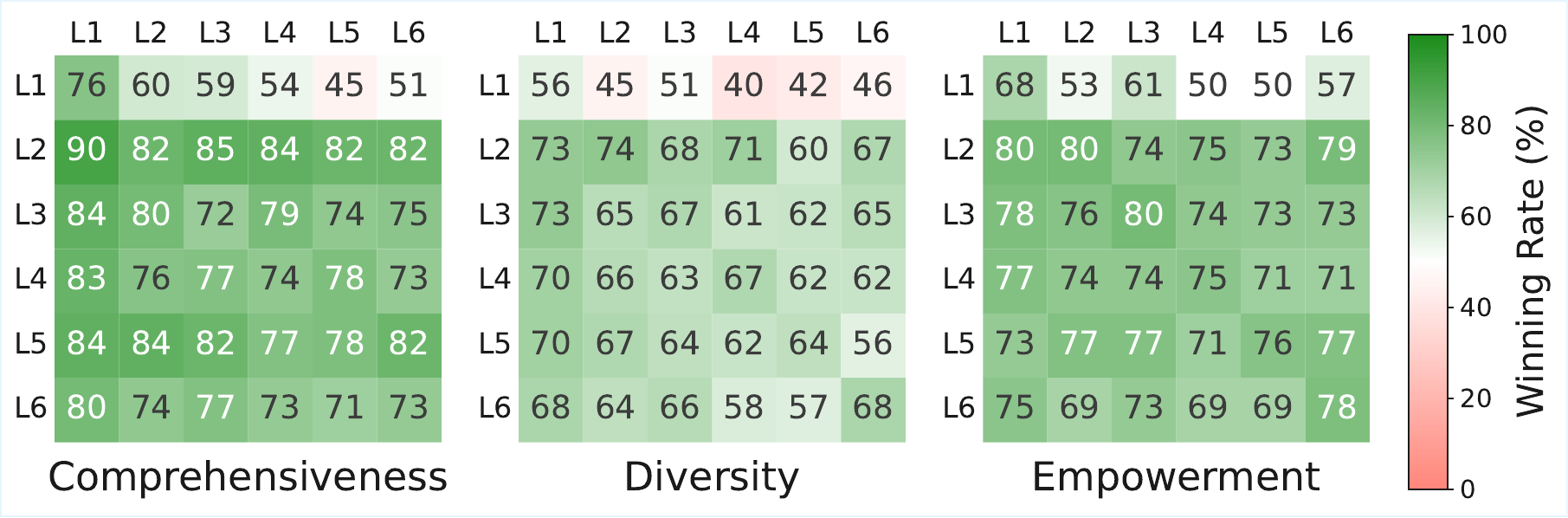}
    \caption{Podcast}
    \label{fig:topic-podcast}
  \end{subfigure}
  
  \vspace{0.5em}  
  \begin{subfigure}[t]{\linewidth}
    \centering
    \includegraphics[width=\linewidth]{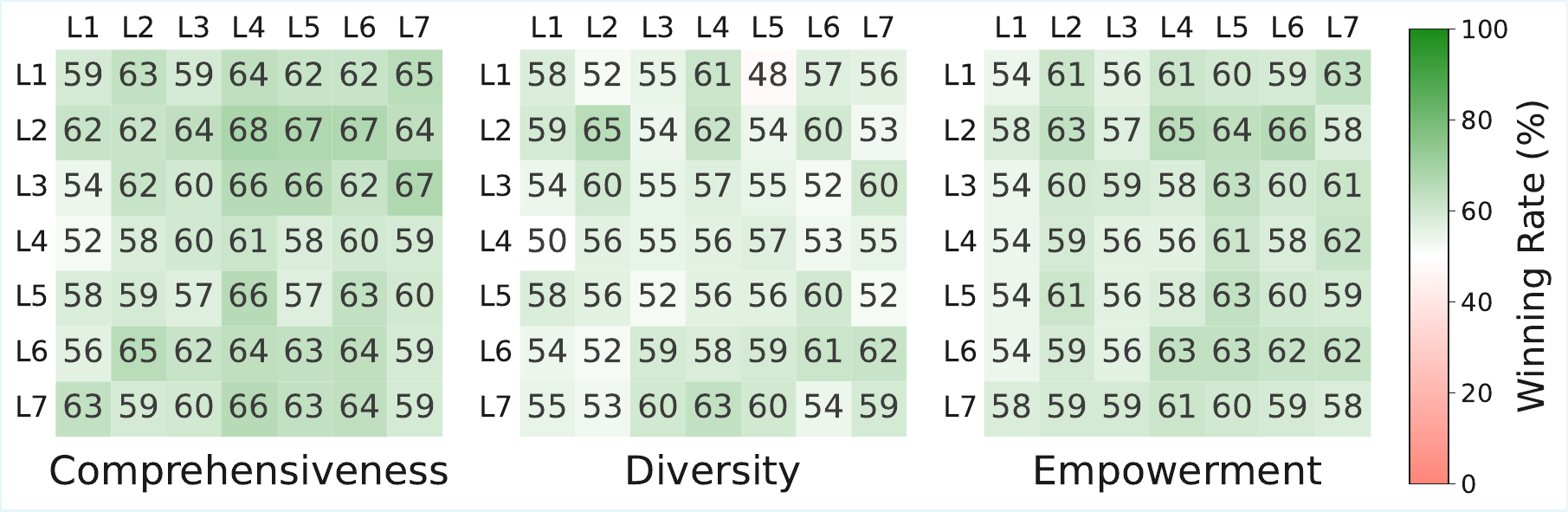}
    \caption{News Articles}
    \label{fig:topic-news}
  \end{subfigure}
  \caption{\textbf{Cross-Level Winning Rates between Models with and without Topic Augmentation.}
  We measured the winning rates (\%) of the model with topic augmentation (rows) over the one without it (column) across community levels in both Podcast (a) and News Articles (b). The results indicate that Topic Augmentation generally performs better than the baseline across different levels.}
\label{fig:topic-augmentation}
\end{figure}

\noindent\textbf{Cross-Level Comparisons} \ \ 
We further compare the outputs of the baseline and those with topic augmentation across different levels, as shown in \cref{fig:topic-augmentation}. 
Note that as the level $l$ increases, the inference complexity generally rises as well, due to the increased number of community summaries.
Therefore, when outcomes are comparable, selecting smaller $l$ is preferable. 
As the results show, topic augmentation not only outperforms the baseline at the corresponding levels but also surpasses the results with larger $l$, requiring more compute.
This suggests that with topic augmentation, we can select smaller $l$, which in turn improves efficiency. 
For example, $l=2$ with topic augmentation produces better results than baseline responses from all levels, while the inference time is just 8 seconds—approximately 2.5 times faster than level 6 of the baseline.

\begin{table}[t]
\centering
\begin{subtable}[t]{\linewidth}
\centering
\scalebox{0.75}{
\begin{tabular}{cccccc}
\toprule
\multicolumn{1}{c}{}
& \multicolumn{3}{c}{Winning Rate (\%) $\uparrow$}
& \multicolumn{2}{c}{Inference Time (sec) $\downarrow$}
\\
\cmidrule(lr){2-4}
\cmidrule(lr){5-6}
Level & Comp. & Div. & Emp. & w/o RA & w/ RA \\
\midrule
1 & 50.0 & 50.0 & 50.0 & 1.49 & 1.98 \\
2 & 78.4 & 59.6 & 65.6 & 8.15 & 7.02 \\
3 & 69.6 & 56.4 & 65.2 & 16.60 & 7.09 \\
4 & 76.4 & 60.0 & 69.6 & 19.09 & 7.06 \\
5 & 79.6 & 64.8 & 67.2 & 19.27 & 7.07 \\
6 & 78.2 & 64.5 & 74.2 & 19.32 & 7.02 \\
\bottomrule
\end{tabular}
}
\vspace{-0.3em} 
\caption*{(a) Podcast}
\end{subtable}

\vspace{0.5em} 

\begin{subtable}[t]{\linewidth}
\centering
\scalebox{0.75}{
\begin{tabular}{cccccc}
\toprule
\multicolumn{1}{c}{}
& \multicolumn{3}{c}{Winning Rate (\%) $\uparrow$}
& \multicolumn{2}{c}{Inference Time (sec) $\downarrow$}\\
\cmidrule(lr){2-4}
\cmidrule(lr){5-6}
Level & Comp. & Div. & Emp. & w/o RA & w/ RA \\
\midrule
1 & 50.0 & 50.0 & 50.0 & 2.68 & 3.19 \\
2 & 61.6 & 50.8 & 58.8 & 20.06 & 7.84 \\
3 & 75.2 & 64.4 & 66.8 & 49.56 & 7.69 \\
4 & 70.4 & 65.6 & 69.6 & 60.52 & 7.62 \\
5 & 75.2 & 68.0 & 70.0 & 62.12 & 7.59 \\
6 & 73.2 & 61.2 & 66.0 & 62.36 & 7.56 \\
7 & 75.3 & 66.5 & 65.9 & 62.54 & 7.62 \\
\bottomrule
\end{tabular}
}
\vspace{-0.3em} 
\caption*{(b) News Articles}
\end{subtable}
\caption{\textbf{Effects of Retrieval Augmentation.} We measured the winning rates (\%) and inference time of the model with retrieval augmentation (RA) at the same community level. 
We use topic augmented models.
The results show that with RA is both effective and efficient. }
\label{tab:wo_retrieval}
\end{table}
\subsection{Effects of Retrieval Augmentation}
\noindent\textbf{Performance and Inference Time} \ \ 
To evaluate the effects of retrieval augmentation ($p=200$ documents), we integrate it with the topic-augmented model.
\cref{tab:wo_retrieval} compares the model with and without retrieval augmentation in terms of winning rate and inference time.
Overall, retrieval augmentation yields improvements in performance while significantly reducing inference time by 65.87\% and 38.04\% on average across all levels for News Articles and Podcast, respectively.
The efficiency gains are particularly notable at larger $l$, where the number of summaries increases.
At level 1, ReTAG takes longer because there are fewer than $p=200$ summaries, so it retrieves all summaries and adds an extra LLM prompting for topic classification. 
Note that since the inputs are identical in both settings, the winning rates are all 50\% at level 1. 
The statistical significance of \cref{tab:wo_retrieval} was reported in \cref{app:llm-eval}.

\begin{figure}[t]
  \centering
  \begin{overpic}[width=1.0\linewidth]{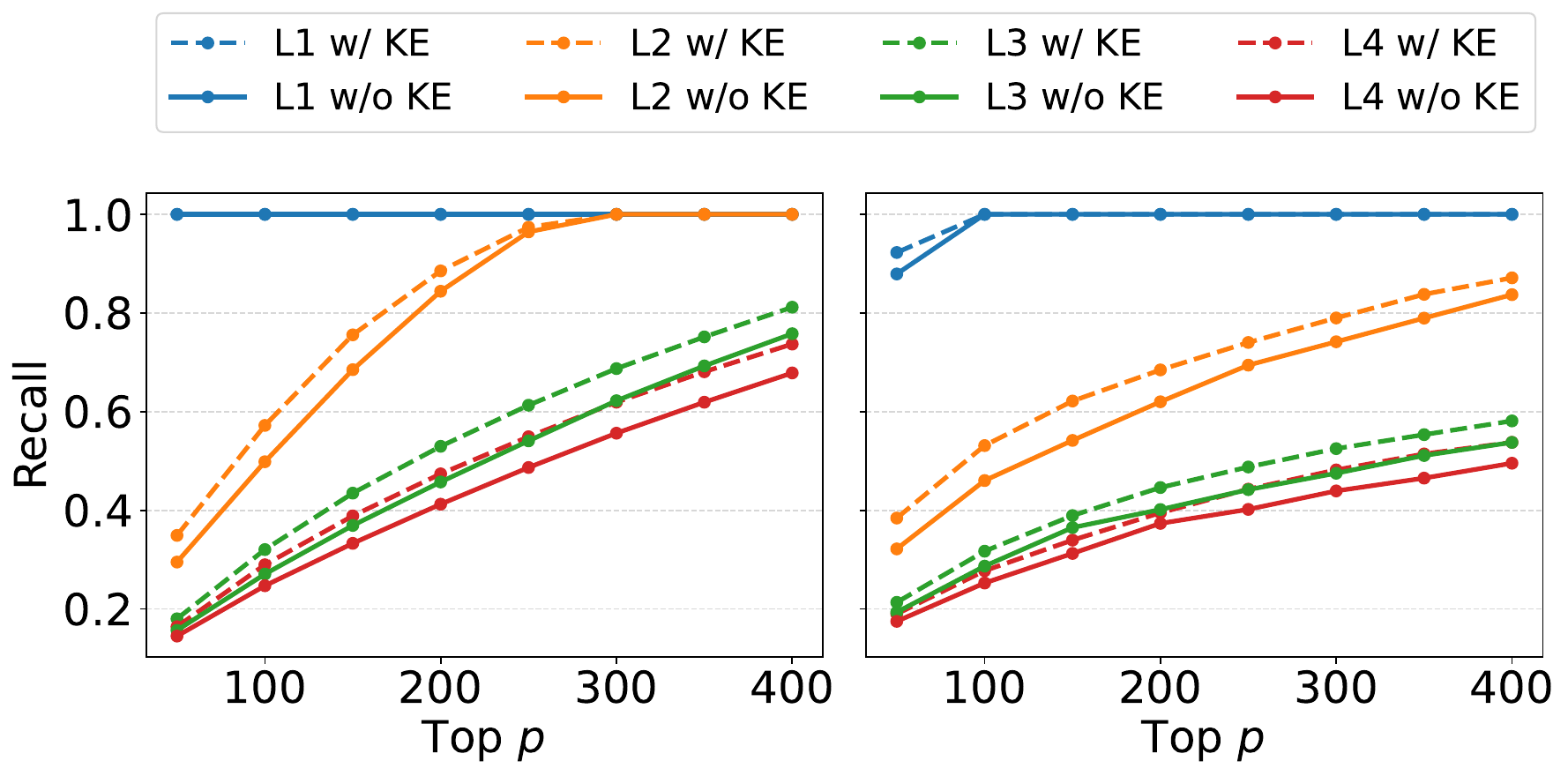}
    \put(21,-3){\small (a) Podcast}
    \put(62,-3){\small (b) News Articles}
  \end{overpic}
  
  \vspace{0.2em}

  \caption{\textbf{Recall@$p$ with Relevant Summaries with and without Keyword Expansion (KE).} 
  Recalls are measured across four levels.
  Relevant summaries are identified by LLM.
  }
  \label{fig:llama-recall}
\end{figure}

\noindent\textbf{Retrieval Performance and Response Generation} \ \ 
We independently evaluate the retrieval performance by measuring recalls using summary-to-query relevance identified by the LLM in \cref{tab:llama-gt-community} as pseudo-GT labels. 
\cref{fig:llama-recall} shows Recall@$p$ with varying $p$ from 50 to 400 for the top 4 levels.
We observe that the recall values quickly approach 1 as $p$ increases up to level 2, ensuring that all relevant documents are used in response generation at these levels. 
In contrast, the recalls with larger $l$ are relatively lower. 
However, the quality of their final responses remains comparable to the baseline scores (\eg, when $p=200$ at larger $l$ in \cref{subsec:retrieval-augmentation}).
We conjecture that this suggests the underlying
\begin{table}[t!]
\centering
\begin{subtable}[t]{\linewidth}
\centering
\scalebox{0.75}{
\begin{tabular}{cccccc}
\toprule
\multicolumn{1}{c}{}
& \multicolumn{3}{c}{Winning Rate (\%) $\uparrow$}
& \multicolumn{2}{c}{Inference Time (sec) $\downarrow$} \\
\cmidrule(lr){2-4}
\cmidrule(lr){5-6}
Level
& Comp.
& Div.
& Emp.
& Baseline
& ReTAG
\\
\midrule
1 & 70.8 & 59.2 & 70.4 & 1.46 & 1.98 \\
2 & 86.4 & 76.8 & 85.2 & 8.71 & 7.02 \\
3 & 82.8 & 74.0 & 83.2 & 17.58 & 7.09 \\
4 & 83.2 & 75.2 & 78.0 & 20.04 & 7.06 \\
5 & 83.6 & 68.4 & 75.6 & 20.33 & 7.07 \\
6 & 85.5 & 75.8 & 80.2 & 20.28 & 7.02 \\
\bottomrule
\end{tabular}
}
\vspace{-0.3em} 
\caption*{(a) Podcast}
\end{subtable}

\vspace{0.5em}

\begin{subtable}[t]{\linewidth}
\centering
\scalebox{0.75}{
\begin{tabular}{cccccc}
\toprule
& \multicolumn{3}{c}{Winning Rate (\%) $\uparrow$}
& \multicolumn{2}{c}{Inference Time (sec) $\downarrow$} \\
\cmidrule(lr){2-4}
\cmidrule(lr){5-6}
Level
& Comp.
& Div.
& Emp.
& Baseline
& ReTAG
\\
\midrule
1 & 55.2 & 54.4 & 54.0 & 2.36 & 3.19 \\
2 & 74.8 & 71.2 & 73.6 & 20.93 & 7.84 \\
3 & 77.2 & 71.6 & 67.2 & 58.64 & 7.69 \\
4 & 76.8 & 75.6 & 71.6 & 75.40 & 7.62 \\
5 & 77.6 & 68.4 & 78.8 & 80.97 & 7.59 \\
6 & 81.2 & 75.2 & 70.8 & 77.96 & 7.56 \\
7 & 82.4 & 73.6 & 76.9 & 77.96 & 7.62 \\
\bottomrule
\end{tabular}
                                                                     }
\vspace{-0.3em} 
\caption*{(b) News Articles}
\end{subtable}
\caption{\textbf{Comparisons between Baseline and ReTAG.} We measured the winning rates (\%) and inference time of
ReTAG at the same community level.
The results show that ReTAG is effective and significantly efficient.}
\label{tab:combined_time}
\end{table}

contextualized entity-relation graph already aggregates related contexts across the corpus, capturing a global sense of the query. 
As a result, each summary contains some degree of relevant global context, and missing a few summaries may not significantly impact the generation of a coherent global response, especially when $l$ is large.

\noindent\textbf{Keyword Expansion} \ \ 
In \cref{fig:llama-recall}, we additionally compare recalls of the retriever with and without keyword expansion technique.
The results show that incorporating keywords into the query enhances retrieval quality resulting in higher recalls. 
This finding highlights the effectiveness of keyword expansion in retrieval augmentation.

\begin{table*}[t]
\centering
\scalebox{0.75}{
\begin{tabular}{lll}
\toprule
\multicolumn{1}{c}{}
& \multicolumn{1}{l}{Baseline}
& \multicolumn{1}{l}{ReTAG} 
\\
\midrule
Question & 	\multicolumn{2}{l}{\makecell[l]{In what ways do sports events influence public health policies, and how is this reflected in the news coverage\\ across the sports and health categories?}}\\
\midrule
Selected Topic & -- & Sports \\
\midrule
\makecell[l]{Community Summary} & \makecell[l]{The World Health Organization and the Centers for \\ Disease Control and Prevention are also key entities \\ in this community, working on issues such as antibiotic \\ resistance and disease tracking…} & \makecell[l]{The CDC and WHO are collaborating to address \\ the issue of antibiotic-resistant bacteria, which \\ poses a significant \textbf{threat to public health}, \\ including in \textbf{sports settings} where infections can\\ spread rapidly…} \\
\midrule
Predicted Answer & \makecell[l]{The influence of sports events on public health policies \\ and their reflection in news coverage across sports and \\ health categories is not directly addressed in the \\ provided reports…} & \makecell[l]{Sports events \textbf{can significantly influence public} \\ \textbf{health policies} in various ways, including \\ promoting physical activity, healthy lifestyles, \\ and disease prevention…} \\
\bottomrule
\end{tabular}
}
\caption{\textbf{Qualitative Examples of ReTAG.} 
We compared qualitative examples from ReTAG with those from the baseline. For the same question, the preserved details and corresponding response predicted by ReTAG were presented in bold. Through this comparison, we confirmed that ReTAG preserves question-relevant details more effectively than the baseline and consequently generates higher-quality responses.}
\label{tab:qualitative-example}
\end{table*}
\subsection{Results of ReTAG}
\cref{tab:combined_time} shows the final performance of ReTAG, which integrates both topic and retrieval augmentation. Across both datasets, ReTAG achieves a winning rate exceeding 50\% in all cases and significantly reduces inference time—\eg, by 90.3\% and 65.4\% at level 6 for the News Articles and Podcast datasets, respectively.
These results demonstrate that ReTAG successfully combines high performance in global sensemaking with rapid inference speed, thanks to the proposed topic and retrieval augmentation techniques. 
The statistical significance of \cref{tab:combined_time} was reported in \cref{app:llm-eval}.

\subsection{Qualitative Analysis}
We present qualitative examples in \cref{tab:qualitative-example} to analyze the performance of ReTAG. Specifically, we report the given query from the News Articles, the corresponding topic selected for the query (a step not performed in the baseline), as well as the summaries and final responses generated by both the baseline and ReTAG.

As shown in \cref{tab:qualitative-example}, when asked about news coverage regarding the influence of sports on public health policy, the baseline failed to reflect such information because the general graph summary did not include the relevant details. In contrast, through Topic Augmentation, ReTAG indexed graph for the Sports topic captured details like the rapid spread of infections in sports settings that threaten public health, enabling the model to generate a more specific and informative response. 
\section{Conclusion}
We introduce ReTAG, a Retrieval-Enhanced, Topic-Augmented Graph-based global sensemaking approach over large document corpora.
By integrating topic-augmented summarization and retrieval-based answer generation with keyword expansion, ReTAG overcomes key limitations of prior graph-based methods.
Our experiments show that ReTAG improves global sensemaking while significantly reducing inference time.

\section*{Limitations}
One limitation shared by ReTAG and the baseline lies in the cost associated with indexing topic-augmented community summaries. 
Constructing and indexing topic-specific subgraphs introduces a non-trivial preprocessing overhead; however, this indexing needs to be done only once per corpus and can be amortized across multiple queries, making it a manageable trade-off for the gains in inference efficiency and response quality. 
And since the indexing depends on using a high-performing LLM to achieve effective global sensemaking, this further increases computational costs due to the resource-intensive nature of such models.

\section*{Ethics Statement}
This study adheres to ethical standards, emphasizing fairness, transparency, and responsibility. All datasets used in this work are publicly available, curated, and free of personally identifiable information. The News Articles dataset is released under the Open Data Commons Attribution License (ODC-By) v1.0, which permits modification, redistribution, and use with proper attribution. The Podcast dataset is provided by Microsoft Corporation and is © 2025 Microsoft Corporation. All rights reserved. It is available for research purposes only, and redistribution, modification, or commercial use is not permitted without explicit permission from Microsoft. The META LLAMA 3.3 model is released under the LLAMA 3.3 COMMUNITY LICENSE AGREEMENT (Release Date: December 6, 2024), which governs responsible use, modification, and redistribution in accordance with Meta’s terms.

\section*{Acknowledgments}
This work was the result of project supported by KT(Korea Telecom)-Korea University AICT R\&D Center.
This research was supported by IITP grants (
IITP-2025-RS-2020-II201819, 
IITP-2025-RS-2024-00436857, 
IITP-2025-RS-2024-00398115, 
IITP-2025-RS-2025-02263754, 
IITP-2025-RS-2025-02304828) and the KOCCA grant
(RS-2024-00345025) funded by the Korea government(MSIT, MOE and MSCT).
\bibliography{custom}
\clearpage
\appendix
\label{sec:appendix}
\section{GraphRAG Framework Details}
\subsection{Contextualized Entity-Relation Graph Construction}\label{appendix:lowest-level-graph}

We provide here the detailed methodology for constructing the contextualized entity-relation graph $G_c$, as mentioned in the main text \cref{subsubsec:contextualized-entity-relation-graph}. Specifically, the entities, relations, and their corresponding contexts are extracted from each document $d \in \mathcal{D}$ using prompt-based LLM inference by ~\cref{prompt:entity-relation}. 
After the initial extraction, a self-reflection step is performed, allowing an additional extraction cycle if the LLM determines re-extraction is necessary. 
We then construct the graph $G_c$ by defining the extracted entities as nodes and the extracted relations as edges, including their corresponding contexts in the graph $G_c$.

\subsection{Community-Based Summarization}\label{appendix:hierarchical-graph-summaries}
Due to the nature of the Hierarchical Leiden algorithm \citep{DBLP:journals/corr/abs-1810-08473}, not all leaf-level communities reside at the same community level. As a result, a community set at a specific level may fail to cover the entire graph $G_c$, leading to the omission of certain subgraphs. This, in turn, can result in an incomplete representation of the entire document set 
$\mathcal{D}$. To address this issue, in cases where a leaf-level community is not at level $L$, we replicate and insert the community into all lower-level community sets (from $l+1$ to $L$), ensuring that each level-wise community set $\mathcal{C}_l$ collectively covers the entire graph $G_c$, and thereby the complete document corpus $\mathcal{D}$.

\noindent\textbf{Leaf-Level Community Summarization}  \ \  For each leaf-level community $c_{L,i}$, the community context is constructed by collecting all contexts linked to the nodes and edges within the subgraph induced by $c_{L,i}$. If the token length of community contexts exceeds the context window size $W$, the context is reduced by removing edges with the lowest prominence. The prominence of an edge $(u,v)$ is defined as the sum of the in-degree and out-degree of its endpoints; that is, the total number of incoming and outgoing edges connected to nodes $u$ and $v$.
Edges are sorted by their prominence in descending order, and the context is truncated by removing edges and their conneted nodes with the lowest prominence until the context fits within $W$. This reduced context is then used to prompt the LLM to generate the summary $\mathcal{S}(c_{L,i})$ (See \cref{appendix:prompts}).

\noindent\textbf{Non-Leaf Community Summarization} \ \ 
For a non-leaf community $c_{l,i}$ at level $l < L$, its context is constructed by aggregating the contexts (node, edge contexts ) of its descendant leaf communities. If the resulting context exceeds the token limit $W$, an iterative reduction procedure is applied to compress the context. Specifically, starting from the leaf level $L$, the descendant community within $c_{l,i}$ with the largest token-length context is identified, and its context is replaced by the corresponding summary $\mathcal{S}(c)$, which is typically shorter. Subsequently, among all contexts including 
$\mathcal{S}(c)$, the community with the longest token length is identified, and its context is replaced with its summary. If the community has already been replaced with 
$\mathcal{S}(c)$, it is further replaced with the summary of its parent community, and the contexts of its sibling communities are removed to reduce the overall length. This process is repeated until the total context length fits within $W$.

\subsection{Response Generation}
\label{appendix:response-generation}
First, the summary segment of each community summary $\mathcal{S}(c_{l,i})$ is divided into 600-token chunks. Subsequently, the entire chunk set is randomly shuffled, and then reassembled into groups matching the context window size 4,000 to prevent the aggregation of essential contextual information within a single context.
We then generate a sub-answer for each resulting context window by \cref{prompt:response}. 
During the sub-answer generation stage, we request that the LLM assign each context window a helpfulness score reflecting its contribution to answer generation.

Any sub-answer with a zero helpfulness score is discarded; among the remaining sub-answers, we then select the highest-scoring ones in descending order until their cumulative token count does not exceed the context window size $W$. 
These selected sub-answers, together with the query $q$, are then provided to the LLM(\cref{prompt:response}) to generate the final global sensemaking response $a$.
\subsection{Global Sensemaking Query Generation}\label{appendix:query}
We generate global sensemaking queries by providing the \texttt{GPT-4o} API with the general descriptions found on the corpus’s website or in its publications and with explanations of graph-based global sensemaking, then creating five hypothetical users who require global sensemaking responses, five tasks per user, and five questions per task. Since \citet{edge2025localglobalgraphrag} did not release this query data, we re-implemented the procedure.

\section{Analysis of Topics}
\label{appendix:topic-qualitative-analysis}
\begin{table}[t]
\centering
{
\scalebox{0.75}
{
\begin{tabular}{c}
\toprule
\makecell[c]{News Articles (29)}\\
\midrule
\makecell[l]{Gaming, Health and Medicine, Tech, Film, Film industry,} \\
\makecell[l]{Politics, Education, Business, Global Health, Entertainment,} \\
\makecell[l]{Music Industry, Artificial Intelligence, Sports, Environment,} \\
\makecell[l]{Politics and Government, Football, Baseball, Basketball,} \\
\makecell[l]{Gaming Industry, Technology, Education and Research,} \\
\makecell[l]{Music, Entertainment Industry, Crypto and Web3, Health,} \\
\makecell[l]{Finance, Entertainment and Media, Business and Finance} \\
\midrule
Podcast (38) \\
\midrule
\makecell[l]{Natural Language Processing, Biotechnology, Psychology,}\\
\makecell[l]{Autonomous Vehicles, Genetic Engineering, Cybersecurity,}\\
\makecell[l]{Sustainability, Software Development, Entrepreneurship,}\\
\makecell[l]{Mathematics, Energy, Coding Skills, Game Development, }\\
\makecell[l]{Virtual Reality, Education, Environmental Issues, Science,}\\
\makecell[l]{Philosophy, Environmental Sustainability, Neuroscience,}\\
\makecell[l]{Quantum Computing, Data Science, Artificial Intelligence,}\\
\makecell[l]{Music, Robotics, Science Fiction, Healthcare, Technology,}\\
\makecell[l]{Space Exploration, Computer Science, Machine Learning, }\\
\makecell[l]{Networking, Innovation, Neural Networks, Climate Change, }\\
\makecell[l]{Gaming, Environmental Science, Diversity and Inclusion, }\\
\bottomrule 
\end{tabular}}
}
\caption{ \textbf{Topic Mined.} 
Topics extracted from News Articles and Podcast using topic mining.
}\label{tab:topic-num}
\end{table}
\noindent\textbf{Qualitative Analysis of Mined Topics} \ \ We qualitatively analyze the extracted topics to better understand the breadth and nature of the content across the datasets. Our goal is to evaluate whether the extracted topics align with the expected diversity and thematic distribution of the underlying data, which includes both news articles and podcast transcripts.

As shown in~\cref{tab:topic-num}, the extracted topics span a wide array of domains, reflecting the multidisciplinary scope of the source materials. For the \textbf{News Articles}, the topics range from global issues such as \textit{Global Health}, \textit{Politics and Government}, and \textit{Environment}, to industry-specific areas like \textit{Crypto and Web3}, \textit{Film Industry}, and \textit{Business and Finance}. This indicates a balanced representation of both societal challenges and sector-specific developments.

On the other hand, the \textbf{Podcast} topics exhibit a stronger tilt toward scientific and technological discourse. Examples include technical fields such as \textit{Quantum Computing}, \textit{Natural Language Processing}, \textit{Cybersecurity}, and \textit{Robotics}, as well as broader academic and societal themes like \textit{Philosophy}, \textit{Environmental Sustainability}, and \textit{Entrepreneurship}. The presence of topics like \textit{Diversity and Inclusion} and \textit{Psychology} also suggests engagement with contemporary social issues.

Overall, the qualitative spread of topics confirms that the topic modeling approach successfully captured a representative and meaningful range of subject matter, validating its effectiveness for further downstream analysis or categorization tasks.
Additionally, due to the difference in domain coverage, we observed that Podcasts-being largely confined to the technology domain-serve as a more appropriate corpus for global sensemaking.

\noindent\textbf{Appropriateness of Selected Topics} \ \ 
We conducted a human evaluation to assess whether the topics selected during the topic classification process described in \cref{subsubsec:topic-augmented-response-generation} were appropriate for the given queries. Specifically, human annotators rated the appropriateness of the selected topics for all questions in News Articles on a 5-point scale. The evaluation yielded an average score of 4.76, indicating that the selected topics were highly appropriate.

\section{Analysis of Graph Indexing Time}
\begin{table}[t]
\centering
\scalebox{0.75}{
\begin{tabular}{lcc}
\toprule
\multicolumn{1}{c}{}
& \multicolumn{1}{c}{Podcast}
& \multicolumn{1}{c}{News Articles} 
\\
\midrule
\makecell[l]{Baseline Graph  \\ Indexing Time (sec)} & 5555.32 & 14322.35 \\
\midrule
\makecell[l]{Additional Graph Indexing Time \\ with Topic Augmentation (sec)} & 178561.94 & 317092.25 \\
\midrule
\makecell[l]{The Number of Topics \\ in Topic Augmentation} & 38 & 29\\
\midrule
\makecell[l]{Baseline Inference Time \\ at Level 6 (sec)} & 20.28 & 77.96 \\
\midrule
\makecell[l]{ReTAG Inference Time \\ at Level 6 (sec)} & 7.02 & 7.56 \\
\bottomrule
\end{tabular}
}
\caption{\textbf{Graph Indexing Time.} 
We compared the graph indexing times between topic augmentation and the baseline. }
\label{tab:graph-indexing-time}
\end{table}
We also report in \cref{tab:graph-indexing-time} a comparison of the indexing times between the topic-augmented graph described in \cref{subsec:topic-augmentation} and the baseline graph. The measurements were conducted using two NVIDIA H100 GPUs. While the indexing time of Topic Augmentation increases linearly with the number of mined topics, this cost is incurred only once and is subsequently amortized across queries. Importantly, topic augmentation improves both response quality and inference efficiency (\cref{tab:topic_graph_main,fig:topic-augmentation}). Furthermore, when retrieval augmentation is integrated into the full model, inference time is substantially reduced—by up to 90.3\% for News Articles and 65.4\% for Podcasts (\cref{tab:combined_time}). This effectively transforms an impractically slow baseline into a practical system, thereby eliminating a critical bottleneck and demonstrating a significant advancement.

\section{Comparison with LighGraphRAG}
We also report the comparison results with an additional baseline, LightGraphRAG \citep{guo2024lightrag} using Podcast. To compare it with ReTAG, we re-implemented LightGraphRAG with LLaMA 3.3 70B Instruct FP8 and BM25. In this evaluation, a single LightGraphRAG prediction is compared with the predictions of ReTAG at each level. As shown in \cref{tab:light-rag}, LightGraphRAG has the advantage of efficiency, as it produces a final answer with a single LLM inference by discarding contexts that exceed the maximum token length. However, it suffers from a critical limitation in that it lacks the global context necessary for effective global sensemaking. This absence substantially degrades the quality of the answers and results in ReTAG achieving an overall winning rate exceeding 50\%. Notably, at Level 1, our method achieves comparable inference time while significantly improving both comprehensiveness (66.0\%) and diversity (61.6\%).
\begin{table}[t]
\centering
\scalebox{0.75}{
\begin{tabular}{cccccc}
\toprule
\multicolumn{1}{c}{}
& \multicolumn{3}{c}{Winning Rate (\%) $\uparrow$}
& \multicolumn{2}{c}{Inference Time $\downarrow$} 
\\
\cmidrule(lr){2-4}
\cmidrule(lr){5-6}
Level & Comp. & Div. & Emp. & LightRAG & ReTAG \\
\midrule
1 & 66.0 & 61.6 & 49.6 & 1.52 & 1.98 \\
2 & 88.8 & 85.6 & 76.4 & 1.52 & 7.02 \\
3 & 86.8 & 82.0 & 70.4 & 1.52 & 7.09 \\
4 & 86.8 & 82.0 & 74.8 & 1.52 & 7.06 \\
5 & 88.0 & 82.0 & 68.4 & 1.52 & 7.07 \\
6 & 92.7 & 84.3 & 68.5 & 1.52 & 7.02 \\
\bottomrule
\end{tabular}
}
\caption{\textbf{Comparison between LightGraphRAG and ReTAG.} 
We compared the winning rates (\%) and inference time of ReTAG against LightGraphRAG. The results show that while LightGraphRAG is efficient, ReTAG demonstrates substantially better effectiveness in global sensemaking.}
\label{tab:light-rag}
\end{table}

\begin{table}[t]
\centering
\scalebox{0.75}{
\begin{tabular}{ccc}
\toprule
\multicolumn{1}{c}{Criteria}
& \multicolumn{1}{c}{Human Eval (\%)}
& \multicolumn{1}{c}{LLM Eval (\%)} 
\\
\midrule
Comprehensiveness & 79 & 82 \\
Diversity & 77 & 81 \\
Empowerment & 75 & 71\\
\bottomrule
\end{tabular}
}
\caption{\textbf{Human and LLM Winning Rates (\%).} 
We conducted a blind human evaluation using randomly selected samples from News Articles. The results show consistent trends between human and LLM evaluation scores. The results show
consistent trends between human and LLM evaluation scores.}
\label{tab:human-eval}
\end{table}
\begin{table*}[!htbp]
\centering
\begin{subtable}[t]{\linewidth}
\centering
\scalebox{0.75}{
\begin{tabular}{cccccccccc}
\toprule
\multicolumn{1}{c}{}
& \multicolumn{3}{c}{Comprehensiveness}
& \multicolumn{3}{c}{Diversity}
& \multicolumn{3}{c}{Empowerment}
\\
\cmidrule(lr){2-4}
\cmidrule(lr){5-7}
\cmidrule(lr){8-10}
Level & Mean & Z-value & p-value & Mean & Z-value & p-value & Mean & Z-value & p-value \\
\midrule
1 & 73.68 & -6.00 & 4.76e-09 & 52.88 & -0.55 & 0.59 & 67.68 & -4.51 & 6.60e-06 \\
2 & 83.92 & -8.16 & 1.99e-15 & 69.60 & -6.06 & 8.33e-09 & 78.16 & -7.13 & 5.81e-12 \\
3 & 77.04 & -6.33 & 9.96e-10 & 64.16 & -4.06 & 1.98e-04 & 75.52 & -6.32 & 1.28e-09 \\
4 & 75.04 & -6.04 & 4.76e-09 & 64.80 & -4.31 & 8.21e-05 & 72.08 & -5.74 & 2.50e-08 \\
5 & 78.48 & -6.94 & 1.89e-11 & 60.96 & -3.29 & 1.97e-03 & 71.84 & -5.76 & 2.50e-08 \\
6 & 73.95 & -5.62 & 1.96e-08 & 62.74 & -3.41 & 1.96e-03 & 73.63 & -5.82 & 2.32e-08 \\
\bottomrule
\end{tabular}
}
\vspace{-0.3em}
\caption*{(a) Podcast}
\end{subtable}

\vspace{0.5em}

\begin{subtable}[t]{\linewidth}
\centering
\scalebox{0.75}{
\begin{tabular}{cccccccccc}
\toprule
\multicolumn{1}{c}{}
& \multicolumn{3}{c}{Comprehensiveness}
& \multicolumn{3}{c}{Diversity}
& \multicolumn{3}{c}{Empowerment}
\\
\cmidrule(lr){2-4}
\cmidrule(lr){5-7}
\cmidrule(lr){8-10}
Level & Mean & Z-value & p-value & Mean & Z-value & p-value & Mean & Z-value & p-value \\
\midrule
1 & 57.28 & -2.23 & 0.09 & 56.00 & -1.83 & 0.34 & 53.36 & -0.71 & 0.48    \\
2 & 63.76 & -3.35 & 5.73e-03 & 60.56 & -3.14 & 0.01 & 62.24 & -2.75 & 0.04 \\
3 & 60.48 & -2.40 & 0.08 & 54.32 & -1.27 & 0.82 & 58.88 & -2.04 & 0.21 \\
4 & 59.60 & -1.86 & 0.13 & 53.76 & -0.77 & 1.00 & 57.44 & -1.69 & 0.31 \\
5 & 57.52 & -1.74 & 0.13 & 54.16 & -0.84 & 1.00 & 58.96 & -1.77 & 0.31 \\
6 & 62.56 & -3.00 & 0.02 & 58.00 & -1.96 & 0.30 & 61.28 & -2.76 & 0.04 \\
7 & 62.09 & -2.27 & 0.09 & 54.07 & -0.68 & 1.00 & 58.79 & -1.35 & 0.35 \\
\bottomrule
\end{tabular}
}
\vspace{-0.3em}
\caption*{(b) News Articles}
\end{subtable}

\caption{\textbf{Statistical Significance of Topic Augmentation.} 
We measured the statistical significance of LLM winning rates (\%) of \cref{tab:topic_graph_main}.} 
\label{tab:stat-sig-ta}
\end{table*}
\begin{table*}[!htbp]
\centering
\begin{subtable}[t]{\linewidth}
\centering
\scalebox{0.75}{
\begin{tabular}{cccccccccc}
\toprule
\multicolumn{1}{c}{}
& \multicolumn{3}{c}{Comprehensiveness}
& \multicolumn{3}{c}{Diversity}
& \multicolumn{3}{c}{Empowerment}
\\
\cmidrule(lr){2-4}
\cmidrule(lr){5-7}
\cmidrule(lr){8-10}
Level & Mean & Z-value & p-value & Mean & Z-value & p-value & Mean & Z-value & p-value \\
\midrule
1 & 50.00 & -- & -- & 50.00 & -- & -- & 50.00 & -- & -- \\
2 & 76.72 & -7.26 & 1.89e-12 & 58.88 & -2.94 & 9.86e-03 & 66.56 & -4.69 & 8.14e-06\\
3 & 69.92 & -5.57 & 2.53e-08 & 59.12 & -2.81 & 9.86e-03 & 61.60 & -3.35 & 8.11e-04 \\
4 & 76.08 & -6.95 & 7.21e-12 & 58.64 & -2.67 & 9.86e-03 & 69.68 & -5.81 & 3.08e-08 \\
5 & 76.00 & -7.25 & 1.89e-12 & 62.96 & -4.00 & 2.56e-04 & 65.12 & -4.04 & 1.06e-04 \\
6 & 76.94 & -7.26 & 1.89e-12 & 63.23 & -4.40 & 5.37e-05 & 70.65 & -5.52 & 1.35e-07 \\
\bottomrule
\end{tabular}
}
\vspace{-0.3em}
\caption*{(a) Podcast}
\end{subtable}

\vspace{0.5em}

\begin{subtable}[t]{\linewidth}
\centering
\scalebox{0.75}{
\begin{tabular}{cccccccccc}
\toprule
\multicolumn{1}{c}{}
& \multicolumn{3}{c}{Comprehensiveness}
& \multicolumn{3}{c}{Diversity}
& \multicolumn{3}{c}{Empowerment}
\\
\cmidrule(lr){2-4}
\cmidrule(lr){5-7}
\cmidrule(lr){8-10}
Level & Mean & Z-value & p-value & Mean & Z-value & p-value & Mean & Z-value & p-value \\
\midrule
1 & 50.00 & -- & -- & 50.00 & -- & -- & 50.00 & -- & -- \\
2 & 63.68 & -3.48 & 5.10e-04 & 53.20 & -1.22 & 0.22 & 57.28 & -2.42 & 0.02 \\
3 & 74.88 & -6.85 & 4.54e-11 & 62.48 & -3.83 & 6.31e-04 & 66.08 & -4.23 & 9.16e-05 \\
4 & 70.64 & -5.57 & 8.72e-08 & 62.24 & -3.67 & 8.16e-04 & 65.36 & -4.19 & 9.16e-05 \\
5 & 75.36 & -6.13 & 4.47e-09 & 66.08 & -4.76 & 1.18e-05 & 69.60 & -5.21 & 1.15e-06 \\
6 & 72.00 & -5.60 & 8.72e-08 & 58.88 & -2.99 & 5.49e-03 & 63.52 & -3.77 & 3.28e-04 \\
7 & 74.29 & -5.55 & 8.72e-08 & 63.08 & -3.71 & 8.16e-04 & 67.25 & -4.53 & 2.92e-05 \\
\bottomrule
\end{tabular}
}
\vspace{-0.3em}
\caption*{(b) News Articles}
\end{subtable}

\caption{\textbf{Statistical Significance of Retrieval Augmentation.} 
We measured the statistical significance of LLM winning rates (\%) of \cref{tab:wo_retrieval}. Overall p-values were below 0.05, indicating statistical significance at the 95\% confidence level. At Level 1, no p-value is reported, as the inputs are identical in both settings.} 
\label{tab:stat-sig-ra}
\end{table*}
\begin{table*}[!htbp]
\centering
\begin{subtable}[t]{\linewidth}
\centering
\scalebox{0.75}{
\begin{tabular}{cccccccccc}
\toprule
\multicolumn{1}{c}{}
& \multicolumn{3}{c}{Comprehensiveness}
& \multicolumn{3}{c}{Diversity}
& \multicolumn{3}{c}{Empowerment}
\\
\cmidrule(lr){2-4}
\cmidrule(lr){5-7}
\cmidrule(lr){8-10}
Level & Mean & Z-value & p-value & Mean & Z-value & p-value & Mean & Z-value & p-value \\
\midrule
1 & 74.08 & -5.82 & 5.74e-09 & 57.92 & -2.17 & 0.03 & 64.96 & -3.64 & 2.68e-04\\
2 & 87.92 & -9.40 & 3.37e-20 & 74.16 & -7.18 & 4.04e-12 & 81.92 & -8.19 & 1.57e-15 \\
3 & 81.92 & -8.24 & 5.03e-16 & 72.32 & -6.85 & 3.80e-11 & 77.44 & -7.21 & 2.77e-12 \\
4 & 82.00 & -8.35 & 2.72e-16 & 72.88 & -6.58 & 1.63e-10 & 75.76 & -6.86 & 2.05e-11 \\
5 & 81.60 & -7.97 & 3.14e-15 & 68.48 & -5.53 & 6.37e-08 & 72.48 & -5.65 & 3.17e-08 \\
6 & 84.68 & -8.59 & 4.32e-17 & 73.31 & -6.60 & 1.63e-10 & 77.26 & -6.98 & 1.15e-11 \\
\bottomrule
\end{tabular}
}
\vspace{-0.3em}
\caption*{(a) Podcast}
\end{subtable}

\vspace{0.5em}

\begin{subtable}[t]{\linewidth}
\centering
\scalebox{0.75}{
\begin{tabular}{cccccccccc}
\toprule
\multicolumn{1}{c}{}
& \multicolumn{3}{c}{Comprehensiveness}
& \multicolumn{3}{c}{Diversity}
& \multicolumn{3}{c}{Empowerment}
\\
\cmidrule(lr){2-4}
\cmidrule(lr){5-7}
\cmidrule(lr){8-10}
Level & Mean & Z-value & p-value & Mean & Z-value & p-value & Mean & Z-value & p-value \\
\midrule
1 & 56.72 & -1.23 & 0.22 & 53.28 & -1.08 & 0.28 & 53.52 & -1.15 & 0.25 \\
2 & 76.72 & -6.31 & 5.55e-10 & 71.92 & -6.19 & 4.19e-09 & 72.80 & -5.03 & 1.46e-06\\
3 & 78.00 & -6.43 & 3.95e-10 & 71.60 & -6.12 & 5.73e-09 & 69.12 & -4.68 & 5.71e-06 \\
4 & 81.28 & -7.13 & 7.06e-12 & 72.32 & -5.83 & 2.81e-08 & 72.64 & -5.48 & 1.67e-07 \\
5 & 79.04 & -6.62 & 1.48e-10 & 68.24 & -5.12 & 6.13e-07 & 76.72 & -6.57 & 3.59e-10 \\
6 & 79.28 & -6.76 & 8.08e-11 & 71.36 & -5.59 & 8.84e-08 & 74.08 & -6.04 & 9.48e-09 \\
7 & 83.85 & -6.65 & 1.42e-10 & 72.31 & -5.48 & 1.28e-07 & 77.69 & -5.66 & 7.52e-08 \\
\bottomrule
\end{tabular}
}
\vspace{-0.3em}
\caption*{(b) News Articles}
\end{subtable}

\caption{\textbf{Statistical Significance of ReTAG.} 
We measured the statistical significance of LLM winning rates (\%) of \cref{tab:combined_time}. Overall p-values were below 0.05, indicating statistical significance at the 95\% confidence level.} 
\label{tab:stat-sig-retag}
\end{table*}

\section{LLM evaluation analysis}\label{app:llm-eval}
As mentioned in prior works \citep{edge2025localglobalgraphrag,NEURIPS2023_91f18a12,es-etal-2024-ragas}, LLM-based evaluation has become the common practice for assessing complex tasks. To further strengthen the validity of our results, we additionally conducted both statistical significance analysis and human evaluation.

\noindent\textbf{Statistical Significance} \ \ 
Following \citet{edge2025localglobalgraphrag}, we measured statistical significance of \cref{tab:topic_graph_main,tab:wo_retrieval,tab:combined_time} and reported in \cref{tab:stat-sig-ta,tab:stat-sig-ra,tab:stat-sig-retag}. Specifically, we conducted each evaluation five times with a fixed answer order, and computed the average score (Mean). We then applied the Wilcoxon signed-rank test, followed by Holm–Bonferroni correction for multiple comparisons. As shown in \cref{tab:stat-sig-ta,tab:stat-sig-ra,tab:stat-sig-retag}, the addition of ReTAG components increases significance, and ultimately, \cref{tab:stat-sig-retag} reports an overall p-value below 0.05, indicating statistical significance.

\noindent\textbf{Human Evaluation} \ \ 
We conducted a blind human evaluation comparing ReTAG’s predictions at Level 7 against the baseline on 50 randomly selected News Articles samples, with the results reported in \cref{tab:human-eval}. Similar to the LLM-based evaluation, human annotators were asked to assess comprehensiveness, diversity, and empowerment using the same scales. The results show consistent trends between human and LLM evaluation scores.

\clearpage
\onecolumn
\section{Prompts}\label{appendix:prompts}
\newtcolorbox[auto counter]{promptbox}[2][]{
  colback=white, colframe=black!80, fonttitle=\bfseries,fontupper=\small, 
  boxrule=0.8pt, arc=2mm, left=4pt, right=4pt, top=4pt, bottom=4pt,
  title={#2}, #1
}

\subsection{Contextualized Entity-Relation Graph Construction}\label{prompt:entity-relation}
\begin{promptbox}{Entity, Relation Extraction Prompt}
\begin{verbatim}
-Goal-
Given a text document that is potentially relevant to this activity and a list of entity types, 
identify all entities of those types from the text and all relationships among the identified entities.
 
-Steps-
1. Identify all entities. For each identified entity, extract the following 
information:
- entity_name: Name of the entity, capitalized
- entity_type: One of the following types: ["organization", "person", "geo", "event"]
- entity_description: Comprehensive description of the entity's attributes and activities 
Format each entity as ("entity"{tuple_delimiter}<entity_name>
{tuple_delimiter}<entity_type>{tuple_delimiter}<entity_description>)
 
2. From the entities identified in step 1, identify all pairs of (source_entity, target_entity) 
that are *clearly related* to each other.
For each pair of related entities, extract the following information:
- source_entity: name of the source entity, as identified in step 1
- target_entity: name of the target entity, as identified in step 1
- relationship_description: explanation as to why you think the source entity and the target entity 
are related to each other
- relationship_strength: a numeric score indicating strength of the relationship between 
the source entity and target entity
Format each relationship as ("relationship"{tuple_delimiter}<source_entity>
{tuple_delimiter}<target_entity>{tuple_delimiter}<relationship_description>
{tuple_delimiter}<relationship_strength>)
 
3. Return output in English as a single list of all the entities and relationships 
identified in steps 1 and 2. Use **{record_delimiter}** as the list delimiter.
 
4. When finished, output {completion_delimiter}
\end{verbatim}
\end{promptbox}

\clearpage
\subsection{Community-Based Summarization}\label{prompt:community-summarization}
\begin{promptbox}{Leaf-level Community Summarization Prompt}
\begin{verbatim}
---Role--
You are an AI assistant that helps a human analyst to perform general information discovery. 
Information discovery is the process of identifying and assessing relevant information associated 
with certain entities (e.g., organizations and 
individuals) within a network.

---Goal--
Write a comprehensive report of a community, given a list of entities that belong to the community 
as well as their relationships and optional associated claims. 
The report will be used to inform decision-makers about information associated with the community 
and their potential impact. The content of this report includes an overview of the community’s 
key entities, their legal compliance, technical capabilities, reputation, and noteworthy claims.

---Report Structure--
The report should include the following sections:
- TITLE: community’s name that represents its key entities - title should be short but specific. 
When possible, include representative named entities in the title.
- SUMMARY: An executive summary of the community’s overall structure, how its entities are related 
to each other, and significant information associated with its entities.
- IMPACT SEVERITY RATING: a float score between 0-10 that represents the severity of IMPACT posed by
entities within the community. IMPACT is the scored importance of a community.
- RATING EXPLANATION: Give a single sentence explanation of the IMPACT severity rating.
- DETAILED FINDINGS: A list of 5-10 key insights about the community. Each insight should have a 
short summary followed by multiple paragraphs of explanatory text grounded according to the grounding 
rules below. Be comprehensive.
Return output as a well-formed JSON-formatted string with the following format:
{{
"title": <report title>,
"summary": <executive summary>,
"rating": <impact severity rating>,
"rating explanation": <rating explanation>,
"findings": [
{{
"summary":<insight 1 summary>,
"explanation": <insight 1 explanation>
}},
{{
"summary":<insight 2 summary>,
"explanation": <insight 2 explanation>
}}
]
}}

---Grounding Rules--
Points supported by data should list their data references as follows:
"This is an example sentence supported by multiple data references 
[Data: <dataset name> (record ids); <dataset name> (record ids)]."
Do not list more than 5 record ids in a single reference. Instead, list the top 5 most relevant record
ids and add "+more" to indicate that there are more.
For example:
"Person X is the owner of Company Y and subject to many allegations of wrongdoing 
[Data: Reports (1), Entities (5, 7); Relationships (23); Claims (7, 2, 34, 64, 46, +more)]."
where 1, 5, 7, 23, 2, 34, 46, and 64 represent the id (not the index) of the relevant data record. 
Do not include information where the supporting evidence for it is not provided.
\end{verbatim}
\end{promptbox}

\clearpage
\begin{promptbox}{Non-leaf Community Summarization Prompt}
\begin{verbatim}
---Role--
You are an AI assistant that helps a human analyst to perform general information discovery. 
Information discovery is the process of identifying and assessing relevant information associated 
with certain entities (e.g., organizations and individuals), relations, communities within a network.

---Goal--
Write a comprehensive report of a community, given a list of entities that belong to the community 
as well as their relationships and community reports
(If the combined entity and relationship descriptions exceed the token limit, 
they should be replaced with the corresponding community-level report summary.). 
The report will be used to inform decision-makers about information associated with the community and 
their potential impact. The content of this report includes an overview of the community’s key entities,
their legal compliance, technical capabilities, reputation, and noteworthy claims.

---Report Structure--
The report should include the following sections:
- TITLE: community’s name that represents its key entities - title should be short but specific. 
When possible, include representative named entities in the title.
- SUMMARY: An executive summary of the community’s overall structure, how its entities are related 
to each other, and significant information associated with its entities.
- IMPACT SEVERITY RATING: a float score between 0-10 that represents the severity of IMPACT posed by 
entities within the community. IMPACT is the scored importance of a community.
- RATING EXPLANATION: Give a single sentence explanation of the IMPACT severity rating.
- DETAILED FINDINGS: A list of 5-10 key insights about the community. Each insight should have a short 
summary followed by paragraphs of explanatory text grounded according to the grounding rules below. 
Be comprehensive.
Return output as a well-formed JSON-formatted string with the following format:
{{
"title": <report title>,
"summary": <executive summary>,
"rating": <impact severity rating>,
"rating explanation": <rating explanation>,
"findings": [
{{
"summary":<insight 1 summary>,
"explanation": <insight 1 explanation>
}},
{{
"summary":<insight 2 summary>,
"explanation": <insight 2 explanation>
}}
]
}}

---Grounding Rules--
Points supported by data should list their data references as follows:
"This is an example sentence supported by multiple data references 
[Data: <dataset name> (record ids); <dataset name> (record ids)]."
Do not list more than 5 record ids in a single reference. Instead, list the top 5 most 
relevant record ids and add "+more" to indicate that there are more.
For example:
"Person X is the owner of Company Y and subject to many allegations of wrongdoing 
[Data: Reports (1), Entities (5, 7); Relationships (23)]."
where 1, 5, 7, 23 represent the id (not the index) of the relevant data record.
Do not include information where the supporting evidence for it is not provided.
\end{verbatim}
\end{promptbox}

\clearpage
\subsection{Response Generation}\label{prompt:response}
\begin{promptbox}{Sub-Answer Generation Prompt}
\begin{verbatim}
---Role--
You are a helpful assistant responding to questions about data in the tables provided.

---Goal--
Generate a response of the target format that responds to the user’s question, 
summarize all relevant information in the input data tables appropriate for the response length 
and format, and incorporate any relevant general knowledge. 
If you don’t  know the answer, just say so. Do not make anything up.
The response shall preserve the original meaning and use of modal verbs such as "shall", "may" or "will".
Points supported by data should list the relevant reports as references as follows:
"This is an example sentence supported by data references [Data: Reports (report ids)]"
Note: the prompts for SS (semantic search) and TS (text summarization) conditions use 
”Sources” in place of ”Reports” above.
Do not list more than 5 record ids in a single reference. Instead, list the top 5 most relevant record 
ids and add "+more" to indicate that there are more.
For example:
"Person X is the owner of Company Y and subject to many allegations of wrongdoing 
[Data: Reports (2, 7, 64, 46, 34, +more)]. He is also CEO of company X [Data: Reports (1, 3)]" 
where 1, 2, 3, 7, 34, 46, and 64 represent the id (not the index) of the relevant data 
report in the provided tables.
Do not include information where the supporting evidence for it is not provided.
At the beginning of your response, generate an integer score between 0-100 that 
indicates how **helpful** is this response in answering the user’s question. 

---Target response format---
{
    "helpfulness_score":An integer between 0 and 100 that represents how helpful the
    response is in answering the user's question,   
    "answer":The full response that directly addresses the user’s question, 
    following all instructions above
}
\end{verbatim} 
\end{promptbox}

\clearpage
\begin{promptbox}{Response Generation Prompt}
\begin{verbatim}
--Role--
You are a helpful assistant responding to questions about a dataset by synthesizing perspectives 
from multiple analysts.

---Goal--
Generate a response of the target format that responds to the user’s question, summarize all the reports 
from multiple analysts who focused on different parts of the dataset, and incorporate any relevant 
general knowledge. 
Note that the analysts’ reports provided below are ranked in the **descending order of helpfulness**.  
If you don’t know the answer, just say so. Do not make anything up.
The final response should remove all irrelevant information from the analysts’ reports and merge 
the cleaned information into a comprehensive answer that provides explanations of all the key points 
and implications appropriate for the response length and format. 
Add sections and commentary to the response as appropriate for the length and format. 
Style the response in markdown.
The response shall preserve the original meaning and use of modal verbs such as "shall", "may" or "will".
The response should also preserve all the data references previously included in the analysts’ reports, 
but do not mention the roles of multiple analysts in the analysis process. 
Do not list more than 5 record ids in a single reference. 
Instead, list the top 5 most relevant record ids and add "+more" to indicate that there are more.
For example:
"Person X is the owner of Company Y and subject to many allegations of wrongdoing 
[Data: Reports (2, 7, 34, 46, 64, +more)]. He is also CEO of company X [Data: Reports (1, 3)]"
where 1, 2, 3, 7, 34, 46, and 64 represent the id (not the index) of the relevant data record. 
Do not include information where the supporting evidence for it is not provided.

---Target response format---
{
    "answer":The full response that directly addresses the user’s question, following 
    all instructions above
}
\end{verbatim}
\end{promptbox}

\clearpage
\subsection{Topic Mining}\label{prompt:topic-mining}
\begin{promptbox}{Sub-Answer Topic Mining Prompt}
\begin{verbatim}
---Role--
You are a helpful assistant who responds to queries regarding data presented in the provided tables.

---Goal---
Your objective is to generate a response that adheres to the target JSON format. 
Your response should extract a list of topics from the community report of a Graph RAG system. 
This system is a graph-based Retrieval Augmented Generation approach designed to facilitate sensemaking
over a large text corpus.

---Instructions---
1. Topic Extraction with Global Sensemaking in Mind: Analyze the community report of the Graph RAG 
system and identify topics of interest. 
Here, "global sensemaking" refers to questions that can only be answered by having a comprehensive 
understanding of the entire data corpus. 
Focus on topics that are relevant for in-depth analysis and require considering the complete corpus.
2. Identify and extract a list of topics from the community report that might interest users of the 
Graph RAG system.
3. Your final response must conform exactly to the following JSON structure:
{
    "helpfulness_score": <an integer between 0 and 100>,
    "answer": [ "topic 1", "topic 2", ... ]
}
4. The integer provided as helpfulness_score should reflect how helpful your response is in addressing 
topic extraction.

***Do not provide any other explanations. Only output the JSON.***
\end{verbatim}
\end{promptbox}

\begin{promptbox}{Response Topic Mining Prompt}
\begin{verbatim}
---Role--
You are a helpful assistant who responds to queries related to topics derived from an overall graph.

-- Goal --
You are given a list of topics that were previously extracted from an entire graph. 
The graph was built for use in a graph-based Retrieval-Augmented Generation (RAG) system, 
designed to support global sensemaking over a large text corpus.

Your task is to extract a list of high-level topics that:
1. Are mentioned multiple times in the provided list.
2. Represent recurring and significant themes across the entire graph.
3. Are likely to be of interest to users interacting with the Graph RAG system.

-- Instructions --
1. Focus on global sensemaking: identify topics that appear repeatedly and contribute to a 
broader understanding of the corpus.
2. Select only the topics that are mentioned multiple times and represent the main themes of the graph.
3. Output a final list of these topics in the following JSON format:
{
    "answer": ["topic 1", "topic 2", ...]
}
**Only return the JSON. Do not include any explanations.**
\end{verbatim}
\end{promptbox}

\clearpage
\subsection{Dataset Description Mining}\label{prompt:dataset-desc}
\begin{promptbox}{Sub-Answer Dataset Description Mining}
\begin{verbatim}
---Role--
You are a helpful assistant who responds to queries regarding data presented in the provided tables.

---Goal---
Your objective is to generate a response that adheres to the target JSON format.
Your response should extract an overall summary of the primary subjects discussed in the dataset as 
reflected by the community reports of a Graph RAG system.
This system is a graph-based Retrieval Augmented Generation approach designed to convert the dataset 
into a graph and form communities to facilitate sensemaking over a large text corpus.

---Instructions---
1. Community-based Dataset Topic Extraction: Analyze the community reports that summarize how the 
dataset has been transformed into a graph with communities, and extract a high-level summary of the main 
topics or subjects that the dataset primarily addresses.
   Here, "overall" refers to a general summary focusing solely on the content topics and subjects, 
   without extracting details about metadata, structure, or other fine-grained elements.
2. Identify and extract a summary that captures what the dataset is mainly about, based on the themes 
emerging from the community reports.
3. Your final response must conform exactly to the following JSON structure:
{
    "helpfulness_score": <an integer between 0 and 100>,
    "answer": <dataset description>
}
4. The integer provided as helpfulness_score should reflect how helpful your response is in addressing 
the dataset description extraction task.

***Do not provide any other explanations. Only output the JSON.***
\end{verbatim}
\end{promptbox}

\begin{promptbox}{Response Dataset Description Mining}
\begin{verbatim}
---Role--
You are a helpful assistant who responds to queries regarding dataset descriptions derived from an 
overall graph.

---Goal---
You should extract an overall dataset description from the list of dataset descriptions that were 
previously extracted from the entire graph.
The dataset description in your response should represent a summary of what the dataset is mainly about, 
based on the recurring content themes identified in the overall graph.

---Instructions---
1. Dataset Description Extraction with Global Sensemaking in Mind:
   Analyze the given list of dataset descriptions from the entire graph, and identify a summary that 
   encapsulates the primary focus of the dataset.
   Here, "global sensemaking" refers to the comprehensive understanding of recurring content themes 
   across the overall graph.
   Focus on extracting an overall description that is significant for in-depth analysis and that 
   represents the main focus of the dataset.
2. Identify and Extract the Dataset Description: From the provided list, select the elements that are 
most frequently emphasized and that appear to represent the overall content focus of the dataset.
3. Your final response must conform exactly to the following JSON structure:
{
    "answer": <dataset description>
}
***Do not provide any other explanations. Only output the JSON.***
\end{verbatim}
\end{promptbox}

\clearpage
\subsection{Topic-Augmented Community Summarization}\label{prompt:topic-community-summarization}
\begin{promptbox}{Topic-Augmented Entity, Relation Extraction Prompt}
\begin{verbatim}
-Goal-
Given a specific topic and a text document, identify only those entities and relationships 
from the text that are clearly relevant to the specified topic.

-Inputs-
- topic: A string describing the subject or theme of interest.
- text: A text document potentially containing entities and relationships.

-Steps-
1. From the text, identify all entities that are relevant to the topic. 
For each relevant entity, extract the following information: 
- entity_name: Name of the entity, capitalized
- entity_type: One of the following types: ["organization", "person", "geo", "event"]
- entity_description: Comprehensive description of the entity's attributes and activities
Format each entity as ("entity"{tuple_delimiter}<entity_name>{tuple_delimiter}
<entity_type>{tuple_delimiter}<entity_description>)

2. From the entities identified in step 1, identify all pairs of (source_entity, target_entity) 
that are *clearly related* to each other within the context of the topic.
For each pair of related entities, extract the following information:
- source_entity: name of the source entity, as identified in step 1
- target_entity: name of the target entity, as identified in step 1
- relationship_description: explanation as to why you think the source entity and the target entity 
are related to each other
- relationship_strength: a numeric score indicating strength of the relationship between 
the source entity and target entity
Format each relationship as ("relationship"{tuple_delimiter}<source_entity>
{tuple_delimiter}<target_entity>{tuple_delimiter}<relationship_description>
{tuple_delimiter}<relationship_strength>)
 
3. Return output in English as a single list of all the entities and relationships identified 
in steps 1 and 2. Use **{record_delimiter}** as the list delimiter.
 
4. When finished, output {completion_delimiter}.

5. If there are no entities or relationships relevant to the topic, output 
{completion_delimiter}.
\end{verbatim}
\end{promptbox}

\clearpage
\begin{promptbox}{Topic-Augmented Leaf-level Community Summarization Prompt}
\begin{verbatim}
---Role--
You are an AI assistant that helps a human analyst to perform general information discovery. 
Information discovery is the process of identifying and assessing relevant information associated 
with certain entities (e.g., organizations and individuals) within a network.

---Goal--
Write a comprehensive report of a community, given a list of entities that belong to the community 
as well as their relationships and optional associated claims. The report will be used to inform 
decision-makers about information associated with the community and their potential impact. 
In addition, ensure that the summary is explicitly dependent on the provided topic, reflecting 
topic-specific aspects alongside the community's details. The content of this report includes an 
overview of the community’s key entities, their legal compliance, technical capabilities, reputation, 
and noteworthy claims.

---Report Structure--
The report should include the following sections:
- TITLE: community’s name that represents its key entities - title should be short but specific. 
When possible, include representative named entities in the title.
- SUMMARY: An executive summary of the community’s overall structure, how its entities are related to 
each other, and significant information associated with its entities.
- IMPACT SEVERITY RATING: a float score between 0-10 that represents the severity of IMPACT posed by
entities within the community. IMPACT is the scored importance of a community.
- RATING EXPLANATION: Give a single sentence explanation of the IMPACT severity rating.
- DETAILED FINDINGS: A list of 5-10 key insights about the community. Each insight should have a short 
summary followed by multiple paragraphs of explanatory text grounded according to the grounding rules 
below. Be comprehensive.
Return output as a well-formed JSON-formatted string with the following format:
{{
"title": <report title>,
"summary": <executive summary>,
"rating": <impact severity rating>,
"rating explanation": <rating explanation>,
"findings": [
{{
"summary":<insight 1 summary>,
"explanation": <insight 1 explanation>
}},
{{
"summary":<insight 2 summary>,
"explanation": <insight 2 explanation>
}}
]
}}

---Grounding Rules--
Points supported by data should list their data references as follows:
"This is an example sentence supported by multiple data references 
[Data: <dataset name> (record ids); <dataset name> (record ids)]."
Do not list more than 5 record ids in a single reference. Instead, list the top 5 most relevant 
record ids and add "+more" to indicate that there are more.
For example:
"Person X is the owner of Company Y and subject to many allegations of wrongdoing 
[Data: Reports (1), Entities (5, 7); Relationships (23); Claims (7, 2, 34, 64, 46, +more)]." 
where 1, 5, 7, 23, 2, 34, 46, and 64 represent the id (not the index) of the relevant data record. 
Do not include information where the supporting evidence for it is not provided.
\end{verbatim}
\end{promptbox}

\clearpage
\begin{promptbox}{Topic-Augmented Non-leaf Community Summarization Prompt}
\begin{verbatim}
---Role--
You are an AI assistant that helps a human analyst to perform general information discovery. Information
discovery is the process of identifying and assessing relevant information associated with certain
entities (e.g., organizations and individuals), relations, communities within a network.

---Goal--
Write a comprehensive report of a community, given a list of entities that belong to the community 
as well as their relationships and community reports(If the combined entity and relationship 
descriptions exceed the token limit, they should be replaced with the corresponding community-level 
report summary.). The report will be used to inform decision-makers about information associated with 
the community and their potential impact. In addition, ensure that the summary is explicitly dependent 
on the provided topic, reflecting topic-specific aspects alongside the community's details. 
The content of this report includes an overview of the community’s key entities, their legal compliance, 
technical capabilities, reputation, and noteworthy claims.

---Report Structure--
The report should include the following sections:
- TITLE: community’s name that represents its key entities - title should be short but specific. When
possible, include representative named entities in the title.
- SUMMARY: An executive summary of the community’s overall structure, how its entities are related 
to each other, and significant information associated with its entities.
- IMPACT SEVERITY RATING: a float score between 0-10 that represents the severity of IMPACT posed by
entities within the community. IMPACT is the scored importance of a community.
- RATING EXPLANATION: Give a single sentence explanation of the IMPACT severity rating.
- DETAILED FINDINGS: A list of 5-10 key insights about the community. Each insight should have a short
summary followed by multiple paragraphs of explanatory text grounded according to the grounding rules
below. Be comprehensive.
Return output as a well-formed JSON-formatted string with the following format:
{{
"title": <report title>,
"summary": <executive summary>,
"rating": <impact severity rating>,
"rating explanation": <rating explanation>,
"findings": [
{{
"summary":<insight 1 summary>,
"explanation": <insight 1 explanation>
}},
{{
"summary":<insight 2 summary>,
"explanation": <insight 2 explanation>
}}
]
}}

---Grounding Rules--
Points supported by data should list their data references as follows:
"This is an example sentence supported by multiple data references 
[Data: <dataset name> (record ids); <dataset name> (record ids)]."
Do not list more than 5 record ids in a single reference. Instead, list the top 5 most relevant record
ids and add "+more" to indicate that there are more.
For example:
"Person X is the owner of Company Y and subject to many allegations of wrongdoing [Data: Reports (1),
Entities (5, 7); Relationships (23)]."
where 1, 5, 7, 23 represent the id (not the index) of the relevant data record.
Do not include information where the supporting evidence for it is not provided.
\end{verbatim}
\end{promptbox}

\clearpage
\subsection{Topic Classification}\label{prompt:topic-classification}
\begin{promptbox}{Topic Classification Prompt}
\begin{verbatim}
You will receive input in a JSON format with the following structure:
{
    "question": "<question text>",
    "topics": ["<topic1>", "<topic2>", "..."],
    "dataset_description": "<dataset description text>"
}

---Task---
1. Topic selection:
Read the "question" and "dataset descripton" and analyze the list of "topics". Choose one topic 
from the provided topics that is most relevant to the question referring to the dataset description.
The chosen topic must match one of the topics in the list exactly (case and punctuation must be 
identical).
2. Output:
If a matching topic is found, return it exactly as provided in the input list.
If none of the topics is related to the question, return "null" as the value.
3. Output Format:
Your output must be in the following JSON format:
{{
    "choosed_topic": "<selected topic or null>"
}}
4. Reference Information:
Note that summarized reports dependent on the chosen topic will later be used to respond to the question. 
The 'dataset_description' field provides information about the dataset corpus used to create those 
summary reports. Your sole task here is to select the relevant topic based on the question and dataset
description; you do not need to generate the report. 
\end{verbatim}
\end{promptbox}

\subsection{Retrieval Augmentation Keyword Expansion}\label{prompt:keyword-expansion}
\begin{promptbox}{Question Keyword Expansion Prompt}
\begin{verbatim}
You will receive input in a JSON format with the following structure:
{
    "question": "<question text>",
    "dataset_description": "<dataset description text>"
}

---Task---
Using only the information provided in the JSON, analyze the `dataset_description` to identify which 
fields, variables, metrics, or concepts are most critical for answering the `question`. 
Then, generate a list of keywords that you would expect to find in a dataset subset summary report 
to successfully answer the question.

Requirements:
1. Refer exclusively to the `dataset_description`—do not assume any external knowledge.
2. Extract concise keywords (one to three words each) representing dataset attributes, column names, 
feature names, or important concepts.
3. Present your output as a JSON array of strings, for example:
   ["keyword_one", "keyword_two", "important_metric", ...]
4. Do not include any additional commentary or explanation—only output the JSON array.
\end{verbatim}
\end{promptbox}

\clearpage
\subsection{Global Sensemaking Queries}\label{prompt:query}
\begin{promptbox}{User Generation Prompt}
\begin{verbatim}
---Role---
You are a helpful assistant generating five personas of hypothetical users of the Graph RAG system.

---Goal---
Given the high-level description of the dataset, generate a bulleted list of 5 personas that represent 
different types of users who would benefit from using the system.
The Graph RAG system is a graph-based RAG approach that enables sensemaking over the entirety of
a large text corpus.

---Additional Instructions---
For each persona, briefly provide their job and key characteristics.
Do not include details about specific tasks they need to perform or reasons why that persona is expected.

Format your response as a list of python strings with the following structure(Do not wrap your response
with json or python block.):
["persona 1", "persona 2", ...]
\end{verbatim}
\end{promptbox}
\begin{promptbox}{Task Generation Prompt}
\begin{verbatim}
---Role---
You are a helpful assistant generating five tasks of a hypothetical persona of the Graph RAG system.

---Goal---
Given the high-level description of the dataset and the design of the Graph RAG system, and a
hypothetical persona of the Graph RAG system, generate a bulleted list of 5 tasks that represent
different types of tasks that the user would do using the system.
The Graph RAG system is a graph-based RAG approach that enables sensemaking over the entirety of
a large text corpus.

---Dataset Description---
{dataset_desc}

---Response Format---
Your response should be formatted as a JSON object with a single key "task".
The value should be a list of strings representing the tasks, for example: {json.dumps({"task": 
["task1", "task2", "task3", "task4", "task5"]})}
Do not wrap your response with any JSON or code block formatting.
\end{verbatim}
\end{promptbox}
\begin{promptbox}{Query Generation Prompt}
\begin{verbatim}
---Role---
You are a helpful assistant generating five high-level questions based on a given hypothetical
persona and their task using the Graph RAG system.

---Goal---
Given the high-level description of the dataset and the design of the Graph RAG system, and a 
hypothetical persona of the Graph RAG system, generate a bulleted list of 5 questions that be able to
assess a broad understanding of the entire corpus rather than requiring retrieval of specific low-level 
facts.
The Graph RAG system is a graph-based RAG approach that enables sensemaking over the entirety of a large 
text corpus.

---Dataset Description---
{dataset_desc}

Your response should be formatted as a JSON Object like examples.(Do not wrap your response with json 
block.)
\end{verbatim}
\end{promptbox}
\clearpage
\subsection{LLM Evaluation}\label{prompt:eval}
\begin{promptbox}{Head-to-head Winning Rate Comparison}
\begin{verbatim}
---Role—
You are a helpful assistant responsible for grading two answers to a question provided by two different 
people.
---Goal—
Given a question and two answers (Answer 1 and Answer 2), assess which answer is better according to 
the following measure:
{batch['criteria_batch'][i]}
Your assessment should include two parts:
- Winner: either 1 (if Answer 1 is better), 2 (if Answer 2 is better), or 0 if they are fundamentally 
similar.
- Reasoning: a short explanation of why you chose the winner with respect to the measure described above.
Format your response as a JSON object with the following structure:
{{"winner": <1, 2, or 0>, "reasoning": "Answer 1 is better because <your reasoning>."}}

{
    "comprehensiveness": "How much detail does the answer provide to cover all the aspects and 
    details of the question? A comprehensive answer should be thorough and complete, without 
    being redundant or irrelevant. For example, if the question is ’What are the benefits and 
    drawbacks of nuclear energy?’, a comprehensive answer would provide both the positive and 
    negative aspects of nuclear energy, such as its efficiency, environmental impact, safety, 
    cost, etc. A comprehensive answer should not leave out any important points or provide
    irrelevant information. For example, an incomplete answer would only provide the benefits
    of nuclear energy without describing the drawbacks, or a redundant answer would repeat the 
    same information multiple times.",
    "diversity": "How varied and rich is the answer in providing different perspectives and 
    insights on the question? A diverse answer should be multi-faceted and multi-dimensional, 
    offering different viewpoints and angles on the question. For example, if the question is 
    ’What are the causes and effects of climate change?’, a diverse answer would provide 
    different causes and effects of climate change, such as greenhouse gas emissions, 
    deforestation, natural disasters, biodiversity loss, etc. A diverse answer should also 
    provide different sources and evidence to support the answer. For example, a single-source 
    answer would only cite one source or evidence, or a biased answer would only provide one 
    perspective or opinion.",
    "empowerment": "How well does the answer help the reader understand and make informed 
    judgements about the topic without being misled or making fallacious assumptions. Evaluate 
    each answer on the quality of answer as it relates to clearly explaining and providing 
    reasoning and sources behind the claims in the answer."
}
You are tasked with evaluating summary reports for a graph-based Retrieval-Augmented Generation (RAG) 
system that:
1. Converts the entire document corpus into a graph and forms communities.
2. Will perform **global sensemaking**, i.e. answer queries by synthesizing information from 
**all** documents in the corpus.
3. Receives input in JSON format with two fields:
   - "question": the user’s query
   - "report": a summary report generated over a subset of the corpus
Your task is to decide whether the provided summary report contains **sufficient** key information from 
the full corpus to support answering the query under the global sensemaking paradigm.
**Output only** “YES” if the report genuinely helps answer the question, or “NO” if it does not.
Do not output anything else—no explanations, no extra tokens.
\end{verbatim}
\end{promptbox}
\end{document}